\documentclass[10pt,journal,compsoc]{IEEEtran}
\usepackage{graphicx,comment}
\usepackage{amsmath,amsfonts,times,epsfig,color,bm}
\usepackage{soul,xargs}
\usepackage{multirow}
\usepackage{wrapfig}
\usepackage{varwidth}
\usepackage{latexsym}
\usepackage{mathtools,nccmath}
\usepackage{caption}
\usepackage{subcaption}
\usepackage{algorithm,algorithmicx,setspace}
\usepackage{algpseudocode}
\usepackage{bm}
\usepackage{fancyhdr}
\usepackage{blindtext}
\ifCLASSOPTIONcompsoc
  \usepackage[nocompress]{cite}
\else
  \usepackage{cite}
\fi

\ifCLASSINFOpdf
\else
\fi
\def\bbtheta{{\mbox{\boldmath $\theta$}}}

\def\bbSig{{\mbox{\boldmath $\Sigma$}}}
\def\bbphi{{\mbox{\boldmath $\phi$}}}
\def\bbPhi{{\mbox{\boldmath $\Phi$}}}
\def \nrf {D }
\def \noisevar {\sigma_{n}^2}

\def\bbtheta{{\mbox{\boldmath $\theta$}}}

\def\bbSig{{\mbox{\boldmath $\Sigma$}}}
\def\bbphi{{\mbox{\boldmath $\phi$}}}
\def\bbPhi{{\mbox{\boldmath $\Phi$}}}

\def \nrf {n_{\text{RF}} }
\def \noisevar {\sigma_{n}^2}

\newcommandx{\thiswillnotshow}[2][1=]{\todo[disable,#1]{#2}}

\algdef{SE}[SUBALG]{Indent}{EndIndent}{}{\algorithmicend\ }%
\algtext*{Indent}
\algtext*{EndIndent}

\renewcommand{\thesection}{\arabic{section}}

\begin{document}
\title{Surrogate modeling for Bayesian optimization beyond a single Gaussian process}
%
%
%
%

\author{Qin~Lu$^*$,~\IEEEmembership{Member,~IEEE,}
        Konstantinos D. ~Polyzos$^*$,~\IEEEmembership{Student Member,~IEEE,} Bingcong Li,~\IEEEmembership{Member,~IEEE,} and~Georgios~B.~Giannakis,~\IEEEmembership{Fellow,~IEEE}
\IEEEcompsocitemizethanks{\IEEEcompsocthanksitem The authors are with Dept. of Electrical and Computer Engineering,
University of Minnesota, Minneapolis, MN 55455. 
E-mails: qlu@umn.edu; polyz003@umn.edu; lixx5599@umn.edu; georgios@umn.edu
\IEEEcompsocthanksitem $*$ The first two authors declare equal contribution.
}
}


\IEEEtitleabstractindextext{%
\begin{abstract}
Bayesian optimization (BO) has well-documented merits for optimizing \textit{black-box} functions with an {\it expensive} evaluation cost. Such functions emerge in applications as diverse as hyperparameter tuning, drug discovery, and robotics. BO hinges on a Bayesian surrogate model to sequentially select query points so as to balance exploration with exploitation of the search space. Most existing works rely on a single Gaussian process (GP) based surrogate model, where the kernel function form is typically {\it preselected} using domain knowledge. To bypass such a design process, this paper leverages an ensemble (E) of GPs to adaptively select the surrogate model fit on-the-fly, yielding a GP mixture posterior with enhanced expressiveness for the sought function.  Acquisition of the next evaluation input using this EGP-based function posterior is then enabled by Thompson sampling (TS) that requires no additional design parameters. To endow function sampling with scalability, random feature-based kernel approximation is leveraged per GP model. The novel EGP-TS readily accommodates parallel operation. To further establish convergence of the proposed EGP-TS to the global optimum, analysis is conducted based on the notion of {\it Bayesian regret} for both sequential and parallel settings. Tests on synthetic functions and real-world applications showcase the merits of the proposed method.
\end{abstract}

\begin{IEEEkeywords}
Bayesian optimization, Gaussian processes,  ensemble learning, Thompson sampling, Bayesian regret analysis
\end{IEEEkeywords}}

\maketitle

\IEEEdisplaynontitleabstractindextext

\IEEEpeerreviewmaketitle

\IEEEraisesectionheading{\section{Introduction}\label{sec:introduction}}
A number of machine learning and artificial intelligence (AI) applications boil down to optimizing an `expensive-to-evaluate' black-box function, including hyperparameter tuning~\cite{snoek2012practical}, drug discovery~\cite{korovina2020chembo}, and policy optimization in robotics~\cite{cully2015robots}. As in hyperparameter tuning, lack of analytic expressions for the objective function and overwhelming evaluation cost discourage grid search, and adoption of gradient-based solvers. To find the global optimum under a limited evaluation budget, Bayesian optimization (BO) offers a principled framework by
leveraging a statistical model to guide the acquisition of query points on-the-fly~\cite{shahriari2015taking,frazier2018tutorial}.

While BO can automate the selection of the best-performing machine learning model along with its optimal hyperparameters, it still necessitates domain-specific expert knowledge to design both the surrogate model and the acquisition function~\cite{snoek2012practical}. In the Gaussian process (GP) based surrogate model, one has to select the kernel type and the corresponding hyperparameters. Also, decision has to be made on the selection from the available acquisition functions, and the associated design parameters if there is any. Minimizing such design efforts so as to automate BO is especially appealing for modern AI tasks. Given that in many setups BO is inherently time-consuming, parallelizing function evaluations to reduce convergence time is also of utmost importance. Further, rigorous analysis is desired to establish convergence of BO algorithms to the global optimum. To address the aforementioned desiderata, the goal of the present work is to develop a BO method that entails the least tuning efforts, accommodates parallel operation, and enjoys convergence guarantees.

\subsection{Related works}
Prior art is outlined next to contextualize our contributions.

\vspace{0.1cm}

\noindent \textbf{Ensemble BO.}
Several choices are available for the surrogate model, acquisition function, and acquisition optimizer for BO~\cite{shahriari2015taking}. Without prior knowledge of the problem at hand, combining the merits of different options can intuitively robustify performance. As pointed out in the 2020 black-box optimization challenge, ensembling methods can empirically boost BO performance for hyperparameter tuning~\cite{turner2021bayesian}. In a broader sense, the ensemble rule has been applied to BO in different contexts, including high-dimensional input~\cite{wang2018batched}, and meta learning~\cite{feurer2018scalable}. 
In the basic BO setup, combining acquisition functions has been explored for a single GP-based surrogate model in a principled way~\cite{hoffman2011portfolio,shahriari2014entropy}. The complementary setting of an ensemble of (GP) surrogate models with a given acquisition function has {\it not} been touched upon.

\vspace{0.1cm}

\noindent \textbf{Thompson sampling (TS) and regret analysis for BO.}
Since its invention by~\cite{thompson1933likelihood}, TS has not received much attention in the bandit community until the past decade that its empirical success~\cite{chapelle2011empirical} and theoretical guarantees~\cite{russo2014learning} have been well documented. In the context of BO, TS has been recently explored under different settings, including high-dimensionality~\cite{mutny2019efficient}, inputs with categorical variables~\cite{nguyen2020bayesian,gopakumar2018algorithmic}, as well as distributed learning~\cite{kandasamy2018parallelised,hernandez2017parallel}. Without additional design parameters, TS is very attractive for automated machine learning. Convergence of TS for BO has been recently established using regret analysis both in the Bayesian~\cite{russo2014learning,kandasamy2018parallelised}, and in the frequentist setting~\cite{chowdhury2017kernelized, vakili2020scalable}. Although TS has been investigated with a mixture prior for linear bandits~\cite{hong2021thompson}, its counterpart in BO with the associated regret analysis has not been studied so far.

\vspace{0.1cm}

\noindent\textbf{Parallel BO.} To reduce convergence time of BO approaches, parallel function evaluations at distributed computing resources is well motivated.
Coupled with upper confidence bound~\cite{desautels2014parallelizing} and expected improvement~\cite{wang2016parallel} based acquisition rules, this parallel operation typically relies on additional hyperparameters or selection rules to ensure the diversity of query points at different locations. On the other hand, TS-based parallel processing necessitates no additional design as in the sequential setting~\cite{hernandez2017parallel}, and enjoys rigorous convergence guarantees ~\cite{kandasamy2018parallelised}. Moreover, parallel BO has also been investigated for input spaces with high dimensions~\cite{wang2018batched} as well as categorical variables~\cite{nguyen2020bayesian}.

\vspace{0.1cm}

\noindent\textbf{Kernel selection for GPs.} Discovery of the form of the kernel function has been considered for conventional GP learning; see, e.g.,~\cite{teng2020scalable,duvenaud2013structure,kim2018scaling, malkomes2016bayesian}. These approaches usually operate in the batch mode and rely on a large number of samples, thus rendering them inapplicable for BO where data are not only acquired online, but also scarce due to the expensive evaluation cost. 
While an online kernel selection scheme has been put forth for prediction-oriented tasks using a candidate of GP models~\cite{lu2020ensemble}, it entails additional design of the acquisition function before being applied to the BO context. 
How to automatically select the kernel function for the GP model in BO is still unexplored. 

\subsection{Contributions}
Relative to the aforementioned previous works, the contributions of this work are summarized in the following four aspects.
\begin{enumerate}
	\item[c1)] Rather than a single GP surrogate model with a preselected kernel function for BO in previous works, an ensemble (E) of GPs is leveraged here to adaptively select the fitted model 
	for the sought function by adjusting the per-GP weight on-the-fly. Capitalizing on the {random feature (RF)} based approximation per GP, acquisition of the next query input is facilitated by TS with scalability and no additional design parameters.
	\item[c2)] The resulting EGP-TS approach readily accommodates parallel function evaluation (a)synchronously.
	\item[c3)] Convergence of the novel EGP-TS approach to the global maximum is established by {\it sublinear} Bayesian regret for both the sequential and parallel settings. 
	\item[c4)] Tests on synthetic functions and real-world applications, including hyperparameter tuning for three machine learning models and robot pushing tasks,  demonstrate the merits of EGP-TS relative to the single GP-based TS, and alternative ensemble approaches.
\end{enumerate}

\noindent {\bf Relation with~\cite{lu2020ensemble}.}
The EGP function model has been considered in our previous work~\cite{lu2020ensemble}  for supervised learning tasks. However, its adaptation to the BO context here is novel and well motivated for the purpose of kernel selection that is important in practice. Coping with limited data in BO, this work differs from~\cite{lu2020ensemble} in the following directions.
\begin{itemize}
    \item[i)] Unlike~\cite{lu2020ensemble} that relies on a large dataset of \emph{passively} labelled samples, the novel EGP-based BO entails extra design of acquisition functions, which select query points \emph{actively}. Two novel EGP-based acquistion functions are devised and tested, namely, EGP-TS and EGP-EI.
    \item[ii)] Although random feature-based approximation has been used also by ~\cite{lu2020ensemble}, it serves a different purpose here. In~\cite{lu2020ensemble}, where the number of samples is large, the RF approximation alleviates the computational complexity of updating the GP model; whereas in the current BO context with limited labelled data, RFs are motivated to conduct function sampling with scalability in the TS-based acquisition function.
    \item[iii)] An extra weight and model \emph{reinitializaiton} is needed each time the kernel hyperparameters are updated using all data acquired (cf. lines 10-15 in Alg.~1 ).  
    \item[iv)] Building on the novel EGP-TS approach, Bayesian regret analysis has been conducted to guarantee convergence to the global optimum. The analysis is novel and nontrivial to deal with the additional challenge brought by the EGP prior (cf. the proof sketch following Theorem~1).
\end{itemize}

\noindent\textbf{Notation.} Scalars are denoted by lowercase, column vectors by bold lowercase, and matrices  by bold uppercase fonts. Superscripts $~^\top$ and $~^{-1}$  denote transpose, and matrix inverse, respectively; while $\mathbf{0}_N$ stands for the $N\times1$ all-zero vector; $\mathbf{I}_{N}$ for the $N\times N$ identity matrix, and $\mathcal{N}(\mathbf{x}; \boldsymbol{\mu}, \mathbf{K})$ for the probability density function (pdf) of a Gaussian random vector $\mathbf{x}$ with mean $\boldsymbol{\mu}$, and covariance $\bf K$.

\section{Preliminaries}
Consider the following optimization problem
\begin{align}
	\mathbf{x}_* = \underset{\mathbf{x}\in \mathcal{X}}{\arg \max} 
	\ \ f(\mathbf{x}) ,\label{eq:BO}
\end{align}
where $\mathcal{X}$ is the feasible set for the $d\times 1$ optimization variable $\mathbf{x}$, and the objective $f(\mathbf{x})$ is {\it black-box} with analytic expression unavailable and is often {\it expensive} to evaluate. This mathematical abstraction characterizes a variety of application domains. When tuning  hyperparameters of machine learning models with $\mathbf{x}$ collecting the hyperparameters, the mapping to the validation accuracy $f(\mathbf{x})$ is not available in closed form, and each evaluation is computationally demanding especially for deep neural networks and large data sizes~\cite{snoek2012practical}. For example, it takes 4 days to train BERT-large on 64 TPUs \cite{devlin2018bert}. The lack of  analytic expression discourages one from leveraging conventional gradient-based solvers to find $\mathbf{x}_*$.  Exhaustive enumeration is also inapplicable given the expensive evaluation cost. Fortunately, BO offers a theoretically elegant solution by judiciously selecting query pairs for a given evaluation budget~\cite{shahriari2015taking,frazier2018tutorial}.

In short, BO relies on a statistical surrogate model to extract information from the evaluated input-output pairs $\mathcal{D}_t:=\{(\mathbf{x}_\tau, y_\tau)\}_{\tau=1}^t$ so as to select the next query input $\mathbf{x}_{t+1}$. Specifically, this procedure is implemented iteratively via two steps, that is: \textbf{s1)} Obtain  $p(f(\mathbf{x})|\mathcal{D}_t)$ based on the surrogate model; and, 
\textbf{s2)} Find $\mathbf{x}_{t+1}\! =\! {\arg\max}_{\mathbf{x}\in\mathcal{X}} \ \alpha (\mathbf{x}|\mathcal{D}_t)$ based on $p(\! f(\mathbf{x})|\mathcal{D}_t)$.  
Here, the so-termed acquisition function $\alpha$, usually available in closed form, is designed to balance {\it exploration} with {\it exploitation} of the search space. There are multiple choices for both the surrogate model and the acquisition function, see, e.g., \cite{shahriari2015taking, frazier2018tutorial}. Next, we will outline the GP based surrogate model, which is the most widely used in BO, and TS for the  acquisition function.

\vspace{0.1cm} 
\subsection{GP-based surrogate model and TS for acquisition} \label{sec:GP-TS}
GPs are established nonparametric Bayesian approaches to learning functions in a sample-efficient manner~\cite{Rasmussen2006gaussian}. This sample efficiency makes it extremely appealing for surrogate modeling in BO when function evaluations are expensive. Specifically, to learn $f(\cdot)$ that links the input $\mathbf{x}_\tau$ with the scalar output $y_\tau$ as $\mathbf{x}_\tau \rightarrow f(\mathbf{x}_\tau) \rightarrow y_\tau$, a GP prior is assumed on the unknown $f$  as $f\sim \mathcal{GP}(0, \kappa(\mathbf{x},\mathbf{x}'))$, where $\kappa(\cdot,\cdot)$ is a positive-definite kernel (covariance) function measuring pairwise similarity of any two inputs. Then, the joint prior pdf of function evaluations $\mathbf{f}_t := [f(\mathbf{x}_1),\ldots, f(\mathbf{x}_t)]^\top$ at inputs $\mathbf{X}_t := \left[\mathbf{x}_1, \ldots, \mathbf{x}_t\right]^\top (\forall t)$ is Gaussian distributed as 
$p(\mathbf{f}_t| \mathbf{X}_t) = \mathcal{N} (\mathbf{f}_t ; {\bf 0}_t, {\bf K}_t)$, where ${\bf K}_t$ is a $t\times t$ covariance matrix whose $(\tau,\tau')$th entry is 
$[{\bf K}_t]_{\tau,\tau'} = {\rm cov} (f(\mathbf{x}_\tau), f(\mathbf{x}_{\tau'})):=\kappa(\mathbf{x}_\tau, \mathbf{x}_{\tau'})$.
The value $f(\mathbf{x}_\tau)$ is linked with the noisy output $y_\tau$ via the per-datum likelihood $p(y_\tau|f(\mathbf{x}_\tau)) = \mathcal{N}(y_\tau;f(\mathbf{x}_\tau),\sigma_n^2)$, where $\sigma_n^2$ is the noise variance. The function posterior pdf after acquiring input-output pairs $\mathcal{D}_t$ is then obtained according to Bayes' rule as~\cite{Rasmussen2006gaussian}
\begin{align}
	p(f(\mathbf{x})|\mathcal{D}_t) = \mathcal{N}(f(\mathbf{x}); \hat{f}_{t}(\mathbf{x}), \sigma_{t}^2(\mathbf{x})),
\end{align}
where the mean and variance are expressed via $\mathbf{k}_{t}(\mathbf{x}) := [\kappa(\mathbf{x}_1, \mathbf{x}) \ldots \kappa(\mathbf{x}_t,  \mathbf{x})]^\top$ and $\mathbf{y}_t:=[y_1 \ldots y_t]^\top$
as 
\begin{subequations}
	\begin{align}	
		\hat{f}_{t}(\mathbf{x}) & = \mathbf{k}_{t}^{\top}(\mathbf{x}) (\mathbf{K}_t + \noisevar
		\mathbf{I}_t)^{-1} \mathbf{y}_t \\
		\sigma_{t}^2(\mathbf{x})& = \kappa(\mathbf{x},\mathbf{x})\! -\! \mathbf{k}_{t}^{\top}(\mathbf{x}) (\mathbf{K}_t\! +\! \noisevar\mathbf{I}_t)^{-1} \mathbf{k}_{t}(\mathbf{x}).
	\end{align}\label{eq:plain_gpp}
\end{subequations}\vspace*{-0.5cm}

With the function posterior pdf at hand, one readily selects the next evaluation point $\mathbf{x}_{t+1}$ using TS, where the function maximizer ${\bf x}_*$ in~\eqref{eq:BO} is viewed as random. Specifically, TS selects the next query point by sampling from the posterior pdf $p(\mathbf{x}_*|\mathcal{D}_t) =\int p(\mathbf{x}_*|f(\mathbf{x}))p(f(\mathbf{x})|\mathcal{D}_t)d f(\mathbf{x})$.
Upon approximating this integral using a sample from the function posterior $p(f(\mathbf{x})|\mathcal{D}_t)$, the next query is found as  
\begin{align}
	\mathbf{x}_{t+1} = \underset{\mathbf{x}\in\mathcal{X}}{\arg\max} \  \tilde{f}_t (\mathbf{x}), \ \  \tilde{f}_t (\mathbf{x})\sim p(f(\mathbf{x})|\mathcal{D}_t)  \;.
\end{align}
This random sampling procedure nicely balances exploration and exploitation.
Implementation of sampling a function from the GP posterior $p(f(\mathbf{x})|\mathcal{D}_t)$ can be realized by discretizing the input space $\mathcal{X}$~\cite{kandasamy2018parallelised}, leveraging the RF based parametric approximant~\cite{rahimi2008random, shahriari2014entropy}, or more recently relying on sparse GP decomposition for efficiency~\cite{wilson2020efficiently}.

Specifically, RF-based approximation leverages the spectral properties of (commonly used) stationary kernels to convert nonparametric GP learning into a parametric one, yielding \cite{quia2010sparse,rahimi2008random}
\begin{align}
&	{\check f} (\mathbf{x}) =  \bbphi_{\mathbf{v}}^\top (\mathbf{x}) \bbtheta, \quad \bbtheta\sim \mathcal{N}(\bbtheta; \mathbf{0}_{2\nrf}, \sigma_\theta^2\mathbf{I}_{2\nrf}) \label{eq:f_check}\\
 &\bbphi_{\mathbf{v}} (\mathbf{x})  
	\!:=\!\! \frac{1}{\sqrt{\nrf}}\!\!\left[\sin(\mathbf{v}_1^\top\! \mathbf{x}), \cos(\mathbf{v}_1^\top \!\mathbf{x}), \ldots, \sin(\mathbf{v}_{\nrf}^\top \mathbf{x}), \cos(\mathbf{v}_{\nrf}^\top \mathbf{x})\right]^{\!\top}\!\!\!\!\! ,\nonumber
\end{align}
where $\{{\bf v}_i\}_{i=1}^D$ are drawn i.i.d. from $\pi_{\bar{\kappa}} (\mathbf{v})$ -- kernel $\kappa$'s normalized spectral density , and $\sigma_\theta^2$ is the magnitude of $\kappa$ (cf.~App.~A in the supplementary file).

Henceforth, the function posterior pdf will be captured by $p(\bbtheta|\mathcal{D}_t) = \mathcal{N}(\bbtheta; \hat{\bbtheta}_t, \bbSig_t)$, based on which TS will select the next query point as
 \begin{align}
  &\mathbf{x}_{t+1} = \underset{\mathbf{x}\in\mathcal{X}}{\arg\max} \  \bbphi_{\mathbf{v}}^\top (\mathbf{x})  \tilde{\bbtheta}_t , \ \  \tilde{\bbtheta}_t\sim p(\bbtheta|\mathcal{D}_t) \ .
 \end{align}
It is worth mentioning that the mean $\hat{\bbtheta}_t$ and covariance matrix $\bbSig_t$ can be updated efficiently in a recursive Bayes manner with the inclusion of each new (input, evaluation) pair.

\section{Ensemble GPs with TS for BO}
The performance of BO approaches depends critically on the chosen surrogate model. While most existing works rely on a single GP with {\it preselected} kernel form, we here leverage an ensemble (E) of $M$ GPs, each relying on a kernel function selected from a given dictionary $\mathcal{K}:=\{\kappa^1, \ldots, \kappa^M\}$. Set $\mathcal{K}$ can be constructed with kernels of different types and different hyperparameters. Specifically, each GP $m\in \mathcal{M}:=\{1,\ldots,M\}$ places a unique prior on $f$ as $f|m \sim \mathcal{GP}(0, \kappa^m (\mathbf{x}, \mathbf{x}'))$. Taking a weighted combination of the individual GP priors, yields the EGP prior of $f(\mathbf{x})$ given by
\vspace*{-0.2cm}
\begin{align}
	f(\mathbf{x})\sim \sum_{m=1}^M w^m_0 {\cal GP}(0,\kappa^m (\mathbf{x},\mathbf{x}')),  \;\; \sum_{m=1}^M w^m_0 =1,  \label{eq:EGP_prior}
\end{align}
where $w^m_0:={\rm Pr} (i=m)$ is the prior probability that assesses the contribution of GP model $m$. Here, the {\it latent} variable $i$ is introduced to indicate the contribution from GP $m$. While this non-Gaussian EGP prior~\eqref{eq:EGP_prior} has been advocated for conventional prediction-oriented tasks in~\cite{lu2020ensemble}, the novelty here is its adaptation for BO along with the extra design step needed for query selection. 
Besides EGP for BO, we will employ TS-based acquisition function, which again, relies on sampling from $p(f(\mathbf{x})|\mathcal{D}_t)$. Coupled with the EGP prior~\eqref{eq:EGP_prior}, this posterior pdf is expressed via the sum-product rule as
\vspace*{-0.25cm}
\begin{align}
	{p}(f(\mathbf{x})|\mathcal{D}_{t}) \! &=\! \sum_{m = 1}^M\! {\rm Pr}(i\!=\! m|\mathcal{D}_{t}) {p}(f(\mathbf{x})|  i\!=\! m,\mathcal{D}_{t} ), \label{eq:EGP_post}
\end{align}
which is a mixture of posterior GPs with per-GP weight $w_t^m:={\rm Pr}(i\!=\! m|\mathcal{D}_{t})$ given by 
\begin{align}
	w_t^m \propto {\rm Pr}(i\!=\! m) p(\mathcal{D}_{t}|i\!=\! m) = w_0^m p(\mathcal{D}_{t}|i\!=\! m), \label{eq:w_t}
\end{align}
where $p(\mathcal{D}_{t}|i\!=\! m)$ is the marginal likelihood of the acquired data ${\cal D}_t$ for GP $m$.  
As with sampling from a Gaussian mixture (GM) distribution, drawing a sample $\tilde{f}_t (\mathbf{x})$ from~\eqref{eq:EGP_post} is implemented by the following two steps
\begin{align}
	{m}_t \sim {\cal CAT}(\mathcal{M},\mathbf{w}_t), \;\;\;
	\tilde{f}_t (\mathbf{x})\sim {p}(f(\mathbf{x})|i= m_t ,\mathcal{D}_{t} ),\label{eq:func_samp}
\end{align}
where ${\cal CAT}(\mathcal{M},\mathbf{w}_t)$ represents a categorical distribution that assigns one of the values from $\mathcal{M}$ with probabilities $\mathbf{w}_t:=[w_t^1,\ldots, w_t^M]^\top$.

There are several choices for the function sampling step~\eqref{eq:func_samp} in the novel EGP-TS as mentioned in Sec.~\ref{sec:GP-TS}. Here, we will adopt the random feature (RF) based method since it can not only efficiently draw the function path $\tilde{f}_t (\mathbf{x})$ that is differentiable with respect to ${\bf x}$, but also accommodate incremental updates of $w_t^m$~\eqref{eq:w_t} and ${p}(f(\mathbf{x})|i=m ,\mathcal{D}_{t} )$ across iterates, as elaborated next.


\subsection{RF-based EGP-TS}
When the kernels in the dictionary are shift-invariant, the RF vector $\bbphi_{\mathbf{v}}^m(\mathbf{x})$ per GP $m$ can be formed via~\eqref{eq:phi_x} by first drawing i.i.d. random vectors $\{\mathbf{v}_j^m\}_{j=1}^{\nrf}$ from $\pi_{\bar{\kappa}}^m (\mathbf{v})$, which is the spectral density of the standardized kernel $\bar{\kappa}^m$. Let  $\sigma_{\theta^m}^2$ be the kernel magnitude so that $\kappa^m = \sigma_{\theta^m}^2\bar{\kappa}^m$. The generative model for the sought function and the noisy output $y$  per GP $m$ can be characterized through the $2D\times 1$ vector $\bbtheta^m$ as
\begin{align}
	{p}(\bbtheta^m) &= \mathcal{N} (\bbtheta^m; \mathbf{0}_{2\nrf}, \sigma_{\theta^m}^2\mathbf{I}_{2\nrf})  \nonumber\\
	p(f(\mathbf{x}_t)|i=m,\bbtheta^m )& = \delta (f(\mathbf{x}_t)-\bbphi_{\mathbf{v}}^{m\top} (\mathbf{x}_t)\bbtheta^m)\nonumber\\
	p(y_t|\bbtheta^m, \mathbf{x}_t) &=
	\mathcal{N}(y_t; \bbphi_{\mathbf{v}}^{m\top} (\mathbf{x}_t)\bbtheta^m, \sigma_n^2) \;.\label{eq:LF}
\end{align}
This parametric form readily allows one to capture the function posterior pdf per GP $m$ via $p(\bbtheta^m|\mathcal{D}_t) = \mathcal{N}(\bbtheta^m; \hat{\bbtheta}_t^m, \bbSig_t^m)$, which together with the weight $w_t^m$, approximates the EGP function posterior~\eqref{eq:EGP_post}. Next, we will describe how RF-based EGP-TS selects the next evaluation input $\mathbf{x}_{t+1}$, and propagates the EGP function pdf by updating the set $\{w_t^m, {\bbtheta}_t^m, \bbSig_t^m, m\in\mathcal{M}\}$ from slot to slot.

Given $\mathcal{D}_t$, acquisition of $\mathbf{x}_{t+1}$ is obtained as the maximizer of the RF-based function sample $\tilde{\check{f}}_t (\mathbf{x})$ based on~\eqref{eq:func_samp}, whose detailed implementation is given by
\begin{align}
	 \mathbf{x}_{t+1} &= \underset{\mathbf{x}\in\mathcal{X}}{\arg\max} \ \ \tilde{\check{f}}_t (\mathbf{x}),\ \  \ {\rm where}\    \tilde{\check{f}}_t (\mathbf{x}):=\bbphi_{\mathbf{v}}^{m_t \top} (\mathbf{x}) \tilde{\bbtheta}_t  \nonumber\\
 {m}_t 
&\sim {\cal CAT}(\mathcal{M},\mathbf{w}_t),   \ \tilde{\bbtheta}_t \sim p(\bbtheta^{m_t }|\mathcal{D}_t), \label{eq:EGP_max}
\end{align}
which can be solved using gradient-based solvers because the objective is available in an analytic form.
Upon acquiring the evaluation output $y_{t+1}$ for the selected input $\mathbf{x}_{t+1}$, the updated weight $w_{t+1}^m := {\rm Pr}(i=m|\mathcal{D}_{t},\mathbf{x}_{t+1}, y_{t+1})$ can be obtained per GP $m$ via Bayes' rule as
\begin{align}
	w_{t+1}^m &
	\!	=\! \frac{{\rm Pr}(i\!=\!m|\mathcal{D}_{t},\mathbf{x}_{t+1}\!) {p}(y_{t+1}|\mathbf{x}_{t+1},  \!i\!=\!m,\mathcal{D}_{t})}{{p}(y_{t+1}|\mathbf{x}_{t+1}, \mathcal{D}_{t})} \nonumber\\
 &=\!   \frac{w_t^m \mathcal{N}\!\left(y_{t+1};  \hat{y}_{t+1|t}^{m}, \!(\sigma_{t+1|t}^{m})^2 \right)}{\sum_{m' = 1}^M\! w_t^{m'}\! \mathcal{N}\!\left(y_{t+1};  \!\hat{y}_{t+1|t}^{m'}, \!(\sigma_{t+1|t}^{m'})^2 \!\right)}\label{eq:w_update}, 	\end{align}
where the sum-product rule allows one to obtain the per-GP predictive likelihood as ${p}(y_{t+1}|i\!=\!m,\mathcal{D}_{t}, \mathbf{x}_{t+1})= \int p(y_{t+1}| \bbtheta^m ,\mathbf{x}_{t+1}) 
p(\bbtheta^m|\mathcal{D}_t) d \bbtheta^m  = \mathcal{N}(y_{t+1};\hat{y}_{t+1|t}^{m},(\sigma_{t+1|t}^{m})^2)$
with $\hat{y}_{t+1|t}^{m} =  \bbphi^{m\top}_{\mathbf{v}} (\mathbf{x}_{t+1})\hat{\bbtheta}_t^{m}$ and $(\sigma_{t+1|t}^{m})^2 = \bbphi^{m\top}_{\mathbf{v}}  (\mathbf{x}_{t+1}) \bbSig^m_{t} \bbphi^m_{\mathbf{v}} (\mathbf{x}_{t+1})+\noisevar$.

Further, the posterior pdf of $\bbtheta^m$ can be propagated in the recursive Bayes fashion as
\vspace*{-0.15cm}
\begin{align}
	{p}(\bbtheta^m |\mathcal{D}_{t+1})  & =  	\frac{{p}(\bbtheta^m |\mathcal{D}_{t}\!) {p}(y_{t+1}|\bbtheta^m,\mathbf{x}_{t+1})}
	{{p}(y_{t+1}|\mathbf{x}_{t+1},  i=m,\mathcal{D}_{t})} \nonumber\\
	 & = \mathcal{N}(\bbtheta^m; \hat{\bbtheta}_{t+1}^m, \bbSig^m_{t+1}),\label{eq:theta_up}
\end{align}
where the updated mean $ \hat{\bbtheta}_{t+1}^m$ and covariance matrix $\bbSig^m_{t+1}$ are
\begin{subequations} 
	\hspace*{-0.5cm}
	\begin{align}
		\hspace*{-0.25cm}\hat{\bbtheta}_{t+1}^m &= \hat{\bbtheta}_{t}^m \!+\!  (\sigma_{t+1|t}^{m})^{-2}\bbSig^m_{t} \bbphi^m_{\mathbf{v}}(\mathbf{x}_{t+1})(y_{t+1} - \hat{y}_{t+1|t}^{m}) \\
		\hspace*{-0.25cm}	\bbSig_{t+1}^m &= \bbSig_{t}^m\! \!-\!  (\sigma_{t+1|t}^{m})^{-2}\bbSig^m_{t} \bbphi^m_{\mathbf{v}}\!(\mathbf{x}_{t+1})  \bbphi^{m\top}_{\mathbf{v}}\!\!(\mathbf{x}_{t+1})\bbSig^m_{t}\!. 
	\end{align}
\end{subequations}

\noindent{\bf (Re)initialization.} In accordance with existing BO implementations, EGP-TS initializes with a small number ($t_0$) of evaluation pairs ${\cal D}_{t_0}$ to obtain kernel hyperparameter estimate $\hat{\boldsymbol{\alpha}}_{t_0}^m$ per GP $m$ by maximizing the marginal likelihood. The weight $w_{t_0}^m$ is then obtained via~\eqref{eq:w_t} using $\hat{\boldsymbol{\alpha}}_{t_0}^m$. As proceeding, the kernel hyperparameters per GP are updated every few iterations using all the acquired data, and subsequently the weights are reinitialized via the batch form~\eqref{eq:w_t} using the updated hyperparameters. Between updates of hyperparameters, EGP-TS leverages \eqref{eq:w_update} and \eqref{eq:theta_up} to incrementally propagate the function posterior pdf. Please refer to Alg.~1 for the detailed implementation of (sequential) EGP-TS.

\subsection{Parallel EGP-TS}
As with the single GP-based TS~\cite{kandasamy2018parallelised}, EGP-TS can readily accommodate parallel implementation for both synchronous and asynchronous settings without extra design. Suppose there are $K$ computing centers/workers that conduct function evaluations in parallel. In the {\it synchronous} setup, $K$ query points are assigned for the workers to evaluate simultaneously by implementing~\eqref{eq:func_samp} $K$ times. After all workers obtain the evaluated outputs, the EGP function posterior is then updated using the $K$ input-output pairs. As for the {\it asynchronous} case, whenever a worker finishes her/his job, the EGP posterior  will be updated and the next evaluation point will be acquired.  Note that the asynchronous setup is very similar to the sequential one except that multiple function evaluations are performed at the same time; see Alg.~2 in the supplementary file for details. Alg.~1 contains the implementation of synchronous parallel EGP-TS when $K>1$.

\noindent The following  two remarks are in order.

\noindent \textbf{Remark 1 (EGP with other acquisition functions).} 
Besides TS, the EGP surrogate model can be coupled with other existing single GP-based acquisition functions, including the well-known expected improvement (EI)~\cite{jones1998efficient} and upper confidence bound (UCB)~\cite{srinivas2012information}. The most direct implementation per iteration is to first draw the model index $m_t$ based on the weights ${\bf w}_t$ as in~\eqref{eq:func_samp}, and then proceed with the conventional EI/UCB acquisition rule for GP $m_t$. Results for this preliminary EGP-EI are presented in App.~E. 
Instead of sampling one GP model per iteration, one could alternatively build on the GP mixture pdf to devise the EI or UCB based acquisition rule. Further investigation along this direction is deferred to our future agenda.

\begin{algorithm}[t]
	\caption{EGP-TS }\label{Alg: EGP-TS}
	\begin{algorithmic}[1]
		\State{\textbf{Input:}  Kernel dictionary $\mathcal{K}$, number $D$ of RFs, number $K$ of workers, and $w_0^m = 1/M$ $\forall m$ }
		\newline
		\State{\textbf{Initialization:}  }
		\State Randomly evaluate $t_0$ points to obtain ${\cal D}_{t_0}$;
		\For{$m = 1, 2, \ldots, M$}
		\State Obtain kernel hyperparameters estimates $\hat{\boldsymbol{\alpha}}_{t_0}^m$ by maximizing the marginal likelihood;
		\State Draw $\nrf$ random vectors $\{\mathbf{v}_i^m\}_{i = 1}^{\nrf}$ from  $\pi_{\bar{\kappa}}^m (\mathbf{v})$ using $\hat{\boldsymbol{\alpha}}_{t_0}^m$;
		\State Obtain $w_{t_0}^m$, $\hat{\bbtheta}_{t_0}^m$, and $\bbSig_{t_0}^m$  based on~\eqref{eq:w_t} and~\eqref{eq:batch_parameter};
		\EndFor
		\newline
		\For{$t = t_0, t_0+1, \ldots$}
		\vspace{0.1cm}
		\If{Reinitialization}
		\For{$m = 1, 2, \ldots, M$}
		\State Obtain $\hat{\boldsymbol{\alpha}}_{t}^m$ by marginal likelihood maximization using ${\cal D}_t$;
		\State Draw $\nrf$ random vectors $\{\mathbf{v}_i^m\}_{i = 1}^{\nrf}$ from  $\pi_{\bar{\kappa}}^m (\mathbf{v})$ using $\hat{\boldsymbol{\alpha}}_{t}^m$;
		\State Obtain $w_{t}^m$, $\hat{\bbtheta}_{t}^m$, and $\bbSig_{t}^m$  based on~\eqref{eq:w_t} and~\eqref{eq:batch_parameter};
		\EndFor
		\EndIf
		\For{$k = 1, 2, \ldots, K$}
		\State Sample $m_t^k$ based on pmf $\mathbf{w}_t$;
		\State Sample $\tilde{\bbtheta}_t^k$ from $\mathcal{N}(\hat{\bbtheta}^{m_t^k}_t,\bbSig^{m_t^k}_t)$;
		\vspace{0.1cm}
		\State Obtain $\mathbf{x}_{t+1}^k = \underset{\mathbf{x}\in \mathcal{X}}{\arg \max}  \ \ \tilde{\bbtheta}_t^{k\top}\!\!\bbphi^{m_t^k}(\mathbf{x})$;
		\State Evaluate $\mathbf{x}_{t+1}^k$ to obtain $y_{t+1}^k$;
		\EndFor
		\vspace{0.1cm}
		\State Update $\{w_{t+1}^m,\hat{\bbtheta}_{t+1}^m,\bbSig_{t+1}^m \}_{m}$ with $\{\mathbf{x}_{t+1}^k, y_{t+1}^k\}_{k}$ based on~\eqref{eq:w_update} and ~\eqref{eq:theta_up};
		\State ${\cal D}_{t+1} = {\cal D}_{t}\cup \{\mathbf{x}_{t+1}^k, y_{t+1}^k\}_{k}$;
		\vspace{0.1cm}
		\EndFor
	\end{algorithmic}
\end{algorithm}

\noindent \textbf{Remark 2 (Relation with fully Bayesian GP-based BO).} 
When the dictionary consists of kernel functions of the same type, the EGP prior amounts to a pseudo Bayesian GP model, where the kernel hyperparameters are chosen from a finite set. This EGP-based pseudo Bayesian model achieves a ``sweet spot" between the Bayesian and non-Bayesian treatment of  GP hyperparameters, where the former entails specifying a reasonable prior and also needs demanding MCMC sampling. In addition, the proposed EGP-TS framework not only accommodates different types of kernels, but also enjoys the upcoming convergence guarantees relative to fully Bayesian GP-based BO.

\section{Bayesian regret analysis}\label{sec:regret}
To establish convergence of the proposed EGP-TS algorithm to the global optimum, analysis will be conducted via the notion of Bayesian regret over $T$ slots, that is defined as\vspace*{-0.2cm}
\begin{align}
	& \mathcal{BR}(T):= \sum_{t=1}^T \mathbb{E}[f(\mathbf{x}_*)-f(\mathbf{x}_t)], \label{eq:BR}
\end{align}\vspace*{-0.3cm}\\
where the expectation is over all random quantities, including the function prior, the observations, and the sampling procedure. Unlike previous works that sample the function from a single GP prior~\cite{kandasamy2018parallelised,russo2014learning}, here we draw $f$ from the EGP prior~\eqref{eq:EGP_prior} as
\begin{align}
	m_*\sim {\cal CAT}(\mathcal{M},\mathbf{w}_0), \quad \;\;f(\mathbf{x})\sim {\cal GP}(0,\kappa^{m_*}(\mathbf{x},\mathbf{x'})).\nonumber
\end{align}
This EGP prior presents additional challenge to the regret analysis. Towards addressing this challenge, we will adapt the techniques in~\cite{hong2021thompson}, where TS with a mixture prior is studied for linear bandits, but not in the BO context.

To proceed, we will need the following assumption and intermediate lemmas.

\noindent \textbf{Assumption 1.} (Smoothness of a GP sample path~\cite{ghosal2006posterior}). {\it If $\mathbf{x}\in\mathcal{X}\subset [0,1]^d$ is compact and convex, there exist constants $a,b, L>0$ such that for any $f(\mathbf{x})\sim {\cal GP}(0,\kappa^m (\mathbf{x},\mathbf{x}'))$
	\begin{align}
		{\rm Pr} \left(\sup_{x_j} \left|\frac{\partial f(\mathbf{x})}{\partial x_j} \right|>L \right) \leq ae^{-(L/b)^2},   \forall j\in \{1,\ldots,d\}. \nonumber
\end{align}}

\noindent \textbf{Lemma 1.} (Maximum information gain (MIG)~\cite{srinivas2012information}). {\it  Let $I^m (f;\mathbf{y}_{\cal A})$ represent the Shannon mutual information one can gain about the function $f\sim {\cal GP}(0,\kappa^m)$ using observations $\mathbf{y}_{\cal A}$ evaluated at finite subset $\mathcal{A}:=\{\mathbf{x}_1,\ldots, \mathbf{x}_T\}\subset \mathcal{X}$. For any $m\in {\cal M}$, the MIG for commonly used kernels can be upper bounded by
	\vspace*{-0.2cm}
	\begin{align}
		\gamma_T :=\!\!\!\! \underset{ \substack{\mathcal{A}\subset\mathcal{X}, |\mathcal{A}|=T\\ m\in{\cal M}}}{\max}\! \!\! I^m (f;\mathbf{y}_{\cal A}) \ \ \leq \tilde{\mathcal{O}}(T^c), \ 0\leq c <1 \;,\nonumber  
	\end{align}\vspace*{-0.1cm}\\
	where $\tilde{O}$ ignores polylog factors. }

\noindent \textbf{Lemma~2.}   (Ratio of posterior variances~\cite{desautels2014parallelizing}). {\it Let  $\mathbf{y}_{\cal A}$ and $\mathbf{y}_{\cal B}$ denote the observations when evaluating $f\sim {\cal GP}(0,\kappa^m) $ at $\mathcal{A}$ and $\mathcal{B}$, which are finite subsets of $\mathcal{X}$. With $\sigma^m_{\mathcal{A}}(\mathbf{x})$ and $\sigma^m_{\mathcal{A}\cup\mathcal{B}}(\mathbf{x})$ representing the posterior standard deviation of the GP conditioned on $\mathcal{A}$ and $\mathcal{A}\cup\mathcal{B}$, there exists $\rho_K\geq 1$ so that the following holds for $|\mathcal{B}|<K$
	\begin{align}
		\left(\sigma^m_{\mathcal{A}}(\mathbf{x})\right)^2 \leq \rho_K \left(\sigma^m_{\mathcal{A}\cup\mathcal{B}}(\mathbf{x})\right)^2, \ \ \forall\mathbf{x}\in\mathcal{X}, m\in\mathcal{M}\;.\nonumber
\end{align} }

As stated in~\cite{srinivas2012information}, Assumption~1 is satisfied for various commonly used stationary kernels that are four times differentiable, including Gaussian kernels and Mat{\'e}rn ones with parameter $\nu >2$, which implicitly allows EGP-TS to draw functions with scalability using RFs as in the preceding section. 
The MIG in Lemma~1 plays an important role in the regret bound. It is an information-theoretic measure quantifying the statistical difficulty of BO~\cite{srinivas2012information, russo2014learning}. Lemma~2 will be useful in deriving the regret in the parallel setup. After making these comments, we are ready to present a Bayesian regret upper bound pertinent to EGP-TS in the sequential setting.\\

\begin{figure*}[t]
	\centering
       \includegraphics[width=1\linewidth]{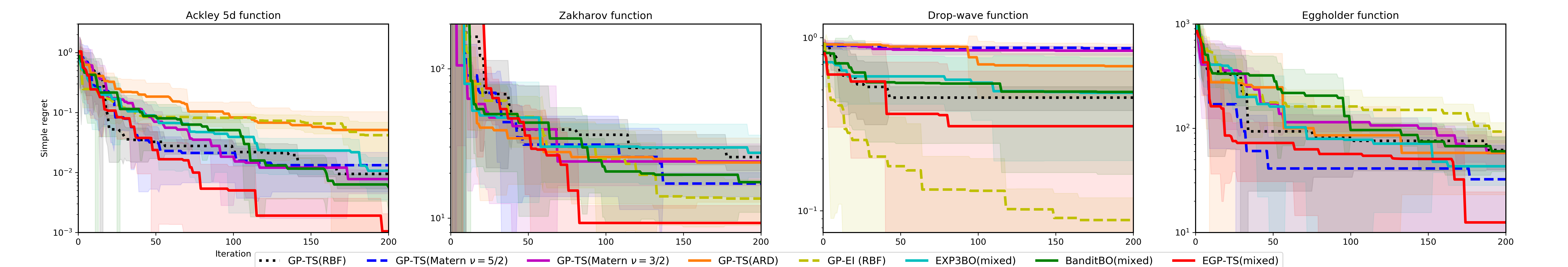}
	\vspace*{-0.55cm}
	\caption{Simple regret on \texttt{Ackley-5d}, \texttt{Zakharov}, \texttt{DropWave} and \texttt{Eggholder} function (from left to right). Dictionary has 4 kernels with distinct forms: RBF with(out) ARD and Mat{\'e}rn with $\nu=3/2, 5/2$. }\label{fig:SR_synFunc_mixed}
\end{figure*}

\begin{figure*}[t]
	\centering
	\includegraphics[width=1\linewidth]{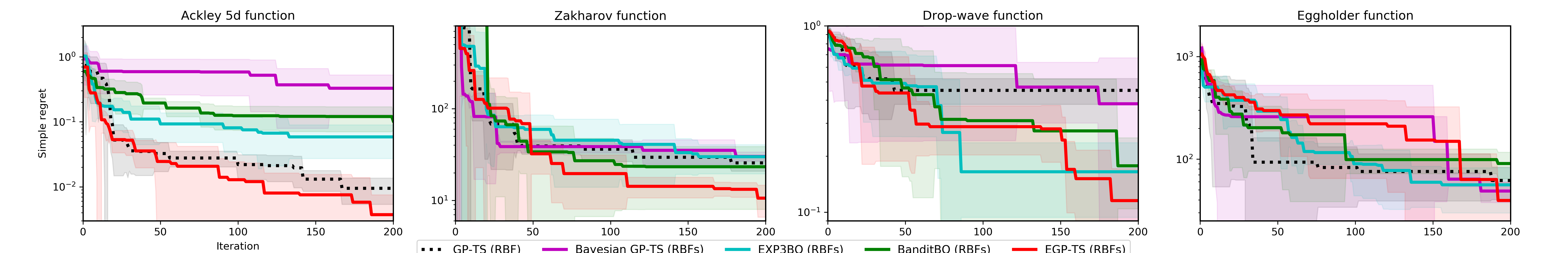}
	\vspace*{-0.55cm}
	\caption{Simple regret on \texttt{Ackley-5d}, \texttt{Zakharov}, \texttt{DropWave} and \texttt{Eggholder} function (from left to right) using RBF kernels. Dictionary has 11 RBF kernels with lengthscales given by $\{10^c\}_{c=-4}^6$.  }\label{fig:SR_synFunc_RBF}
\end{figure*}

\noindent {\bf Theorem 1.} {\it Under Assumption 1, the cumulative Bayesian regret~\eqref{eq:BR} of EGP-TS over $T$ slots, is bounded by
	\begin{align}
		\mathcal{BR}(T) \leq c_1 \sqrt{MT^{c+1}\log T} + 2\sigma_n \sqrt{MT\log T} + c_2, \nonumber
	\end{align}
	where the constants $c_1\!:=\! (2\!+\!\sqrt{d} ) (2/\log (1\!\!+\!\!\sigma_n^{-2})^{1/2}$ and $c_2\!:=\! 9MB \!+\! (\pi^2 d)/6 \!+\! \sqrt{2\pi}M/12$ ($B$ is a constant given in Lemma~3 in App.~B ) are not dependent on $T$}.

\noindent \textbf{Proof sketch.} The detailed proof of Theorem 1 is deferred to App.~B. The key step in the proof builds on the connection with UCB based approaches, that is manifested via decomposing the Bayesian regret~\eqref{eq:BR} as
\begin{align}
	& \mathcal{BR}(T)\!=\underbrace{\sum_{t=1}^T \mathbb{E}[f(\mathbf{x}_*)\!-\!U_t^{m_*}(\mathbf{x}_*)]}_{\mathcal{BR}_1(T)} \!+\!\! \underbrace{\sum_{t=1}^T \mathbb{E}[U_t^{m_t}(\mathbf{x}_t)\!-\!\!f(\mathbf{x}_t)]}_{\mathcal{BR}_2(T)},\nonumber
\end{align}  
where $U^m_t (\mathbf{x}):=\mu_{t-1}^m (\mathbf{x})+\beta_{t}^{1/2} \sigma_{t-1}^m (\mathbf{x})$ with $\beta_{t}$ specified by ~\eqref{eq:beta_t} in App.~B, is a UCB for $f(\mathbf{x})$ under GP $m$. This decomposition of $\mathcal{BR}(T)$ holds since $\{m_t, \mathbf{x}_t\}$ and $\{m_*, \mathbf{x}_*\}$ are i.i.d.  and $U_t^m (\mathbf{x})$ is {\it deterministic} conditioned on $\mathcal{D}_{t-1}$, yielding~\cite{russo2014learning, hong2021thompson},
\begin{align}
	\mathbb{E}_{t-1} [ U_t^{m_t}(\mathbf{x}_t)]=\mathbb{E}_{t-1} [U_t^{m_*}(\mathbf{x}_*)], \ \ \forall t \;. \nonumber
\end{align}
Then, the Bayesian regret bound of EGP-TS can be established by upper bounding ${\cal BR}_1 (T)$ and ${\cal BR}_2 (T)$. Since $f\sim {\cal GP}(0,\kappa^{m_*})$, the former can be conveniently bounded based on related works that rely on a single GP~\cite{russo2014learning, kandasamy2018parallelised}. 
Specifically, ${\cal BR}_1 (T)$ is proved to be upper bounded by a constant, because the probability that $f(\mathbf{x}_*)$ is larger than $U_t^{m_*}(\mathbf{x}_*)$ across all the slots is low~\cite{kandasamy2018parallelised}. 

To further bound ${\cal BR}_2 (T)$ involving the extra latent variable $m_t$ sampled from the EGP posterior (cf.~\eqref{eq:func_samp}), we adapt the technique in~\cite{hong2021thompson} that constructs a confidence set $\mathcal{C}_t$ for the latent variable such that $m_* \in \mathcal{C}_t$ holds with high probability; see Lemma~4 in App.~B. It turns out that ${\cal BR}_2 (T)$ can also be bounded by the sum of posterior standard deviations, which further yields the upper bound given by the MIG along the lines of~\cite{srinivas2012information}.

The proof of Theorem 1 in App.~B involves an additional discretization step of $\mathcal{X}$ per step $t$, in order to cope with the continuous feasible set $\mathcal{X}$.
\\


The following two theorems further establish the cumulative Bayesian regret bounds of parallel EGP-TS in the asynchronous and synchronous settings, whose proofs are deferred to Apps. C-D.


\noindent {\bf Theorem 2 (Asynchronously parallel setting).} {\it For $K$ workers conducting parallel function evaluations asynchronously, EGP-TS under Assumption 1 incurs the following cumulative Bayesian regret over $T$ function evaluations
	\begin{align}
		\mathcal{BR}^{\rm asy}(T) \leq c_1 \sqrt{\!\rho_K MT^{c+1}\log T} \!+\! 2\sigma_n \sqrt{\!\!MT\log T} \!+\! c_2 \nonumber.
\end{align} }


\noindent {\bf Theorem 3 (Synchronously parallel setting).} {\it For $K$ workers performing $T$ function evaluations synchronously, the cumulative Bayesian regret of EGP-TS under Assumption 1 is bounded by
	\begin{align}
		& \mathcal{BR}^{\rm syn}(T) \leq  \ (K\!\!-\!\!1)\sqrt{d\log (K\!\!-\!\!1)}+ 2\sigma_n \sqrt{MT\log T} + c_2
		\nonumber\\
		&+ c_3 \sqrt{\rho_K MT^{c+1}\log T} +  c_4 \sqrt{ MT^{c+1}\log (T+K-1)},  \nonumber
	\end{align}
	where the two constants are given by $c_3:=2 (2/\log (1+\sigma_n^{-2})^{1/2}$, and $c_4:= (2d/\log (1+\sigma_n^{-2})^{1/2}$.
}


The first term of the regret bound in Theorem~2 is $\sqrt{\rho_K}$ times its counterpart in Theorem~1 for the sequential setting. It shall be easily verified that Bayesian regret bounds of parallel EGP-TS become equivalent to that in the sequential setting when $K=1$ with $\rho_1 = 1$. Note that the regret bounds for parallel EGP-TS here are for the number of evaluations, that will typically exceed the bound in the sequential setup. This can be certainly the other way around if the evaluation time is of interest~\cite{kandasamy2018parallelised}. In all the three settings, the cumulative Bayesian regret bounds of EGP-TS  boil down to $\mathcal{O}(\sqrt{MT^{c+1}\log T} )$ after ignoring irrelevant constants, which is {\it sublinear} in the number of evaluations when $0\leq c<1$. Hence, EGP-TS enjoys the diminishing average regret per evaluation as $T$ grows, hereby establishing convergence to the global optimum.




\section{Numerical tests}
In this section, the performance of the proposed EGP-TS will be tested on a set of benchmark synthetic functions, two robotic tasks, and the hyperparameter tuning tasks of three machine learning models. The competing baselines are GP-EI~\cite{jones1998efficient}, the default method for many traditional BO problems, and  TS-based methods, including GP-TS with a preselected kernel type, fully Bayesian GP-TS, as well as two {\it ensemble} approaches, which are BanditBO~\cite{nguyen2020bayesian}, and EXP3BO~\cite{gopakumar2018algorithmic}. It is worth mentioning that the latter two, combining multi-armed bandits and BO, are originally designed for inputs with categorical variables, but are adapted as ensemble methods here with each ``arm" referring to a GP model with the same input variables. 

The kernel hyperparameters per GP for all the TS-based methods other than fully Bayesian GP-TS are obtained by maximizing the marginal likelihood using \texttt{sklearn}.  GP-EI is implemented using \texttt{BoTorch} with the ARD kernel, whose hyperparameters are refitted each iteration.
The fully Bayesian GP model hinges on a pre-defined kernel type where the kernel hyperparameters are assumed to be random variables. In the present work, the RBF kernel is considered and a uniform prior is assumed for the amplitude $\sigma_\theta^2$, characteristic lengthscale and noise variance $\sigma_n^2$, within intervals $[1,100],[10^{-3},10^3]$ and $[0.1,0.3]$ respectively. 
The fully Bayesian GP-TS proceeds by first drawing a sample of the kernel hyperparameters using \texttt{GPyTorch} and \texttt{Pyro} Python packages, based on which function sampling is conducted.
Existing kernel selection methods for conventional GP learning operate in batch mode using a large number of samples, hence being not suitable for the low-data BO setting. For initialization in all the methods, the first $10$ evaluation pairs are randomly selected and used to obtain the kernel hyperarameters per GP by maximizing the marginal likelihood. 
In EGP-TS, the per-GP prior weight is set as uniform, i.e., $w_0^m = 1/M \ \forall m$.
Unless stated otherwise, the hyperparameters are refitted every $50$ iterations for EGP-TS, and every iteration for the rest of the baselines. 
RF approximation with $50$ spectral features is leveraged by all the TS-based approaches for fairness in comparison.
All the experiments are repeated $10$ times, where the average performance and the standard deviation of all competing approaches are reported. 

Additional results concerning ablation studies of the EGP-TS approach, runtime comparison, and the parallel setting are deferred to App.~B in the supplementary file.

\begin{figure*}
	\centering
	\begin{minipage}{.495\textwidth}
		\centering
		\includegraphics[width=0.99\textwidth, height=0.4\textwidth ]{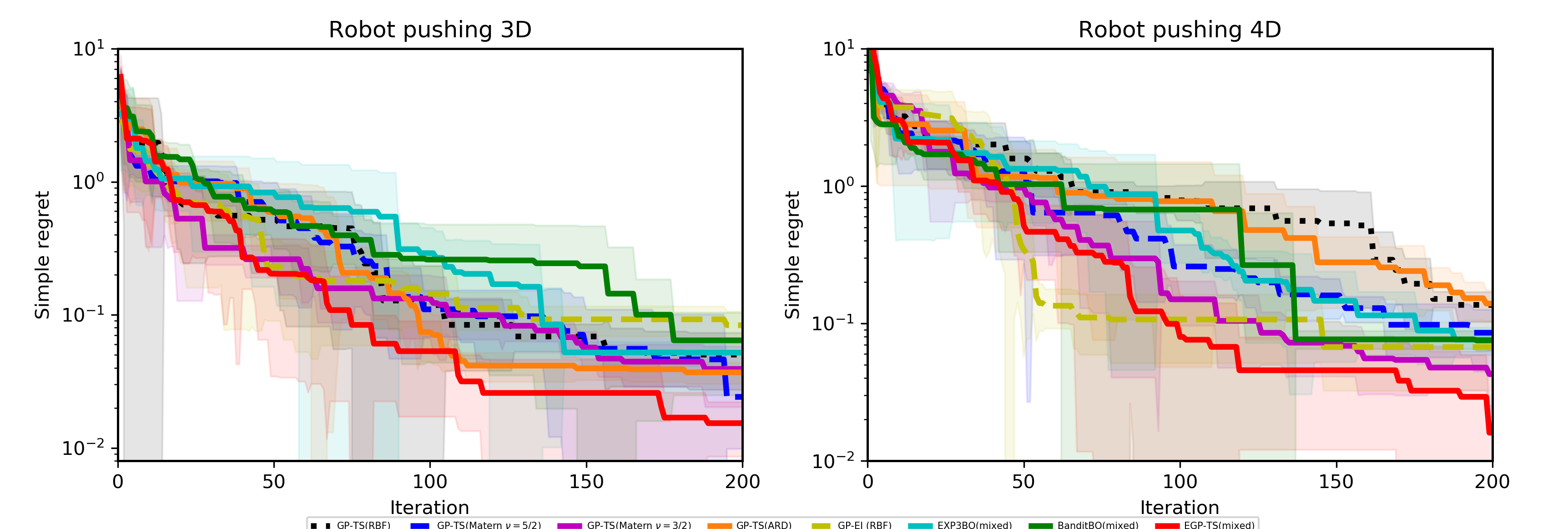}\label{robota}\\{(a)}
	\end{minipage} 
	\begin{minipage}{.495\textwidth}
		\centering
		\includegraphics[width=0.98\textwidth,height=0.4\textwidth]{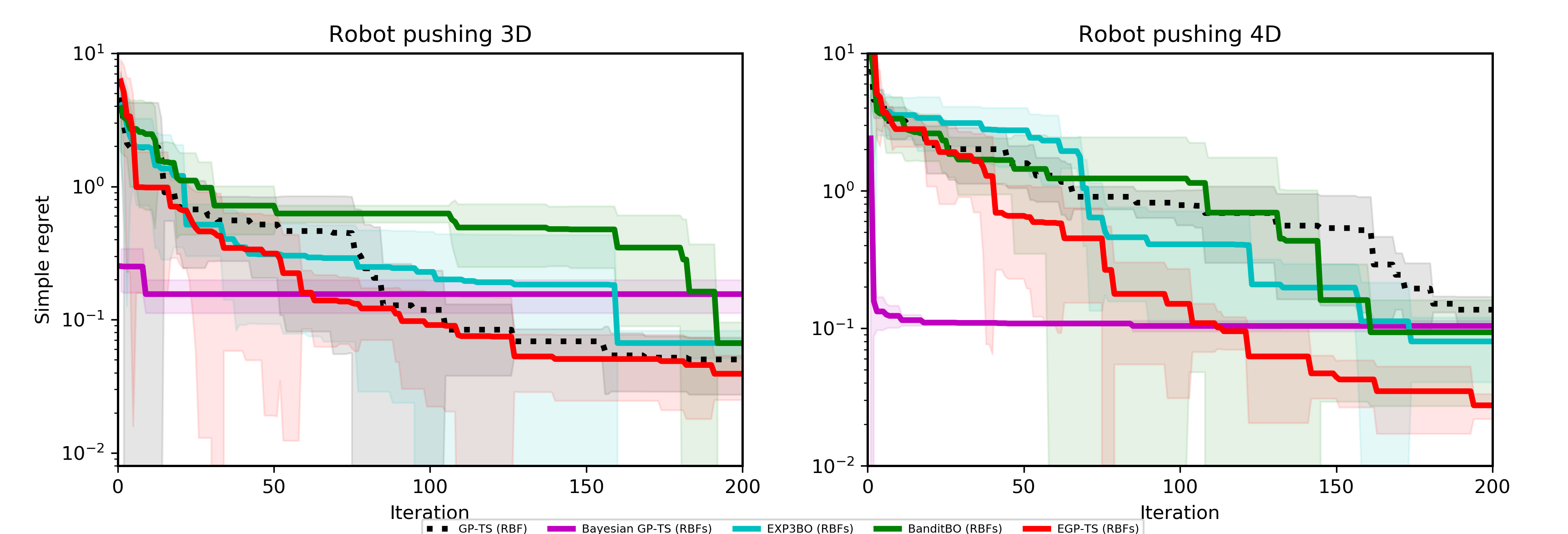}\label{robotb}\\
		{(b)}
	\end{minipage}
	\vspace*{-0.2cm}
	\caption{Simple regret on \texttt{Robot pushing 3D} and \texttt{Robot pushing 4D} tasks with dictionary (a) that has 4 kernels with distinct forms: RBF with(out) ARD and Mat{\'e}rn with $\nu=3/2, 5/2$; and (b) that has 11 RBF kernels with characteristic lengthscales given by $\{10^c\}_{c=-4}^6$.}
	\label{fig:robotplots}
\end{figure*}




\subsection{Tests on synthetic functions} We tested the competing methods on a suite of standard synthethic functions for BO, including \texttt{ Ackley-5d}, \texttt{Zakharov}, \texttt{Drop-wave}, as well as  \texttt{Eggholder}, where the latter two are challenging functions with many local optima.
The performance metric per slot $t$ is given by the simple regret (SR), defined as
${\cal SR} (t) := f(\mathbf{x}_*) - \max_{\tau\in\{1,\ldots,t\}} f(\mathbf{x}_\tau)$. First, to explore the effect of the kernel functions in the (E)GP model, we tested GP-TS with the kernel function being RBF with and without auto-relevance determination (ARD), and Mat{\'e}rn kernels with $\nu=3/2, 5/2$. For all the ensemble methods, the kernel dictionary is comprised of the aforementioned four kernel functions. It is evident from Fig.~1 that the form of kernel function plays an important role in the performance of GP-TS. Combining different kernel functions, EGP-TS not only yields substantially improved performance relative to GP-TS counterparts, but also requires the least design efforts on the choice of the kernel function. In addition, EGP-TS achieves lower simple regret than BanditBO and EXP3BO. Although GP-EI is superior to GP-TS baselines on \texttt{Zakharov} function, EGP-TS yields better performance relative to the former, what demonstrates the benefit of ensembling GP models. 
Upon fixing the kernel type as RBF without ARD and constructing the dictionary as $11$ RBF functions with lengthscales given by $\{10^c\}_{c=-4}^6$, EGP-TS is also compared with fully Bayesian GP-based TS in addition to the aforementioned baselines. Still, EGP-TS outperforms all competitors as shown in Fig.~2. 

\begin{figure*}
	\centering
	\includegraphics[width=0.99\textwidth]{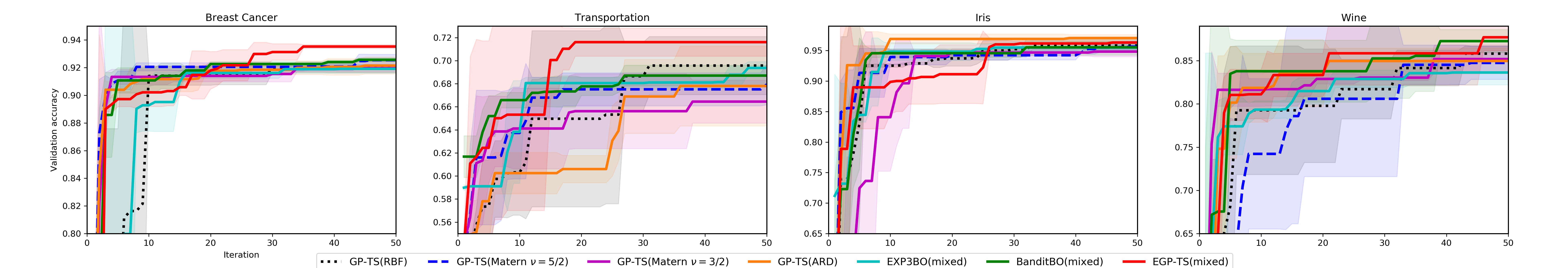}
	\caption{The best validation accuracy (so far) vs. the number of function evaluations on \texttt{Breast Cancer}, \texttt{Transportation}, \texttt{Iris}, and \texttt{Wine} datasets (from left to right) for the NN hyperparameter tuning task. Dictionary has 4 kernels with distinct forms: RBF with(out) ARD and Mat{\'e}rn with $\nu=3/2, 5/2$.}
	\label{fig:NN_mix_hyp}
\end{figure*}

\begin{figure*}
	\centering
	\includegraphics[width=0.99\textwidth]{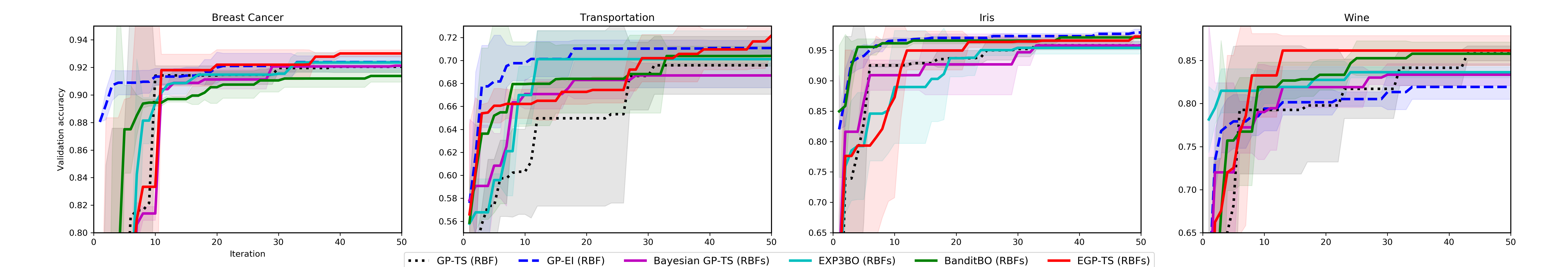}
	\caption{The best validation accuracy (so far) vs. the number of function evaluations on \texttt{Breast Cancer}, \texttt{Transportation}, \texttt{Iris}, and \texttt{Wine} datasets (from left to right) for the NN hyperparameter tuning task. Dictionary has 11 RBF kernels with characteristic lengthscales given by $\{10^c\}_{c=-4}^6$.}
	\label{fig:NN_hyp}
\end{figure*}

\begin{figure*}
	\centering
	\includegraphics[width=0.99\textwidth]{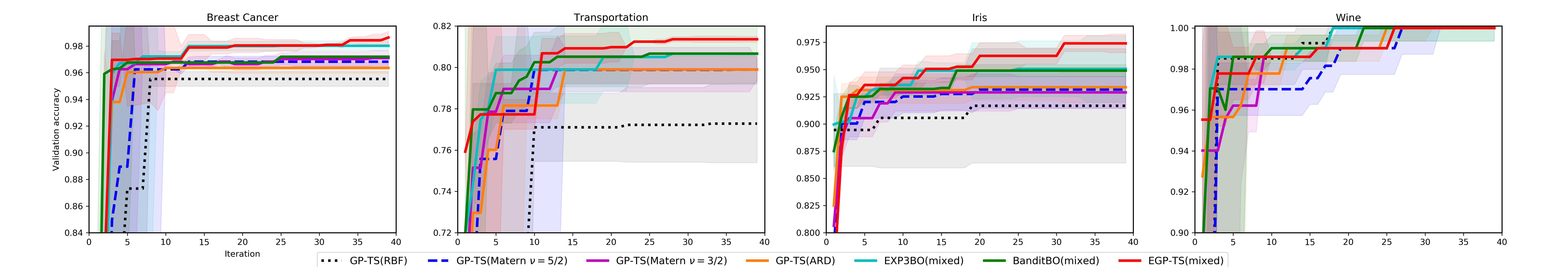}
	\caption{The best validation accuracy (so far) vs. the number of function evaluations on \texttt{Breast Cancer}, \texttt{Transportation}, \texttt{Iris}, and \texttt{Wine} datasets (from left to right) for the SVM hyperparameter tuning task. Dictionary has 4 kernels with distinct forms: RBF with(out) ARD and Mat{\'e}rn with $\nu=3/2, 5/2$.}
	\label{fig:SVM_hyp_mix}
\end{figure*}

\begin{figure*}
	\centering
	\includegraphics[width=0.99\textwidth]{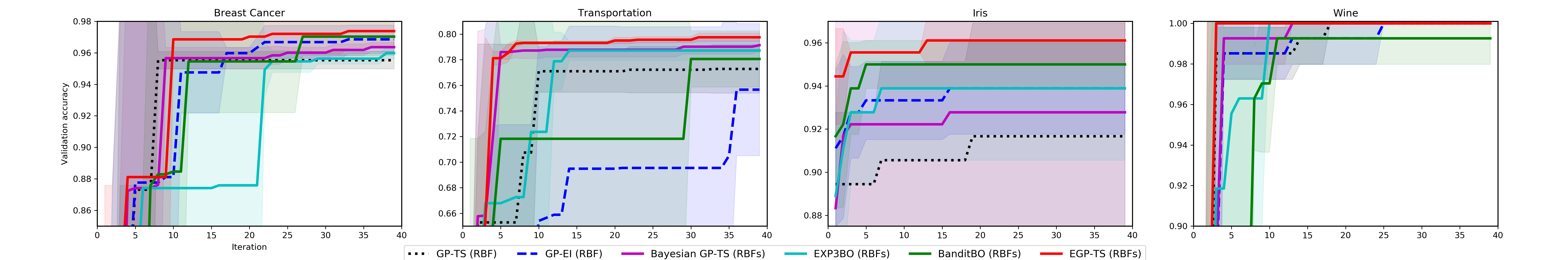}
	\caption{The best validation accuracy (so far) vs. the number of function evaluations on \texttt{Breast Cancer}, \texttt{Transportation}, \texttt{Iris}, and \texttt{Wine} datasets (from left to right) for the SVM hyperparameter tuning task. Dictionary has 11 RBF kernels with characteristic lengthscales given by $\{10^c\}_{c=-4}^6$.}
	\label{fig:SVM_hyp}
\end{figure*}

\begin{figure*}
	\centering
	\includegraphics[width=0.99\textwidth]{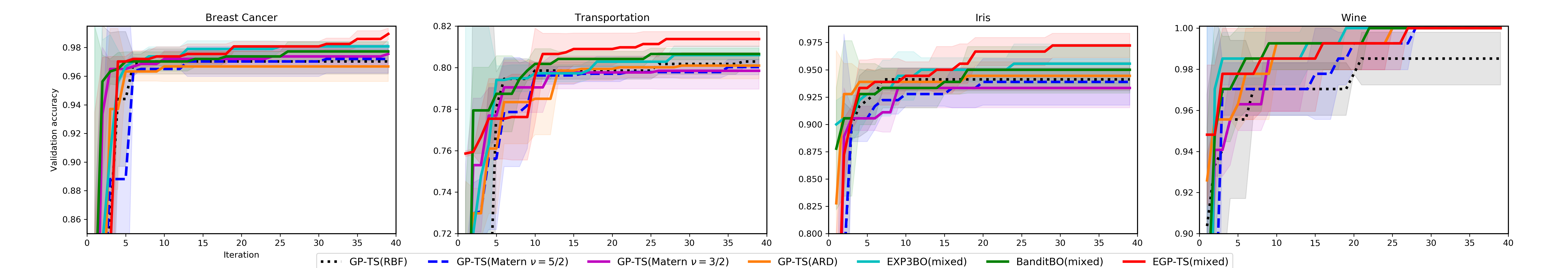}
	\caption{The best validation accuracy (so far) vs. the number of function evaluations on \texttt{Breast Cancer}, \texttt{Transportation}, \texttt{Iris}, and \texttt{Wine} datasets (from left to right) for the GradientBoosting hyperparameter tuning task. Dictionary has 4 kernels with distinct forms: RBF with(out) ARD and Mat{\'e}rn with $\nu=3/2, 5/2$.}
	\label{fig:Grad_hyp_mix}
\end{figure*}

\begin{figure*}
	\centering
    \includegraphics[width=0.99\textwidth]{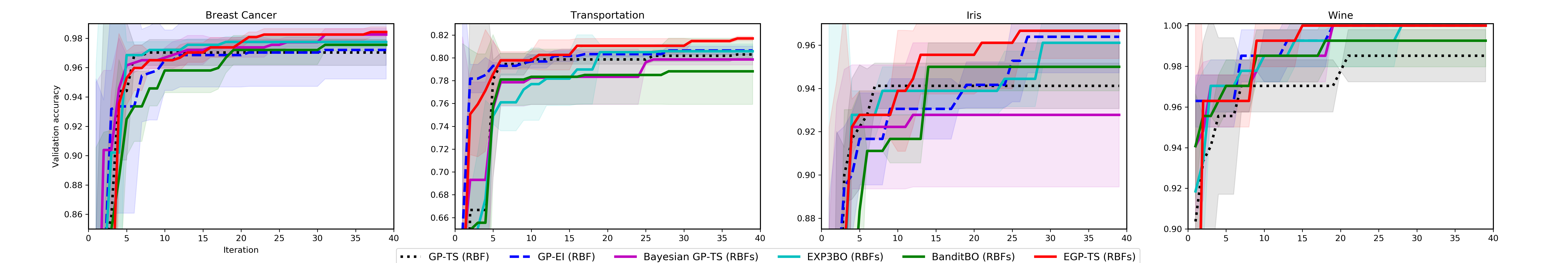}
	\caption{The best validation accuracy (so far) vs. the number of function evaluations on \texttt{Breast Cancer}, \texttt{Transportation}, \texttt{Iris}, and \texttt{Wine} datasets (from left to right) for the GradientBoosting hyperparameter tuning task. Dictionary has 11 RBF kernels with characteristic lengthscales given by $\{10^c\}_{c=-4}^6$.}
	\label{fig:Grad_hyp}
\end{figure*}

\subsection{Robot pushing tasks} 

The second experiment concerns a practical task in robotics, where a robot adjusts its action so as to push an object towards a given goal location. By minimizing the distance between the target location and the end position of the pushed object, we tested two scenarios with $3$ and $4$ input variables following~\cite{wang2017max}. The former optimizes the $2$-D position of the robot and the push duration, and the latter entails optimizing an additional push angle. We used the github codes\footnote{https://github.com/zi-w/Max-value-Entropy-Search} from \cite{wang2017max} to generate the movement of the object pushed by the robot. Each scenario was repeated for $10$ randomly selected goal locations, and the average performance of the competing methods are depicted in Fig.~\ref{fig:robotplots}.
Adaptively selecting kernel function from the dictionary with $4$ distinct forms (that is, RBF with(out) ARD, and Mat{\'e}rn with $\nu=3/2, 5/2$), the proposed EGP-TS outperforms all the competitors, including GP-EI, GP-TS with a preslected kernel, and the other two ensemble methods, as shown in Fig.~\ref{fig:robotplots}(a). The superior performance of EGP-TS when the kernel function is fixed as RBF is also shown in Fig.~\ref{fig:robotplots}(b), what is in accordance with Fig.~2.
It is worth highlighting that EGP-TS not only outperforms fully Bayesian GP-TS in simple regret, but also runs much faster.

\begin{figure*}[t]
	\centering
    \includegraphics[width=1\linewidth]{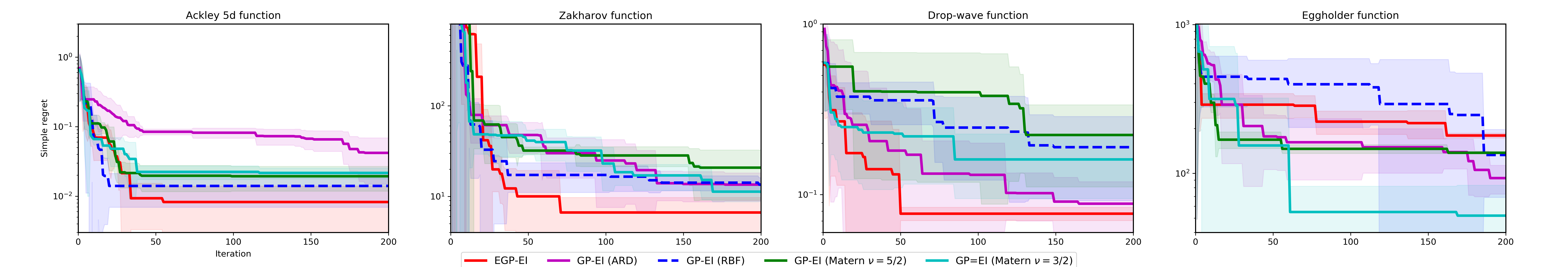}
	\caption{Simple regret of EGP-EI vs. GP-EI with a preselected kernel on \texttt{Ackley-5d}, \texttt{Zakharov}, \texttt{DropWave} and \texttt{Eggholder} function (from left to right). Dictionary has 4 kernels with distinct forms: RBF with(out) ARD and Mat{\'e}rn with $\nu=3/2, 5/2$. }\label{fig:EGP_EI_synth}
\end{figure*}

\begin{figure*}[t]
	\centering
    \includegraphics[width=0.85\linewidth, height = 0.3\linewidth]{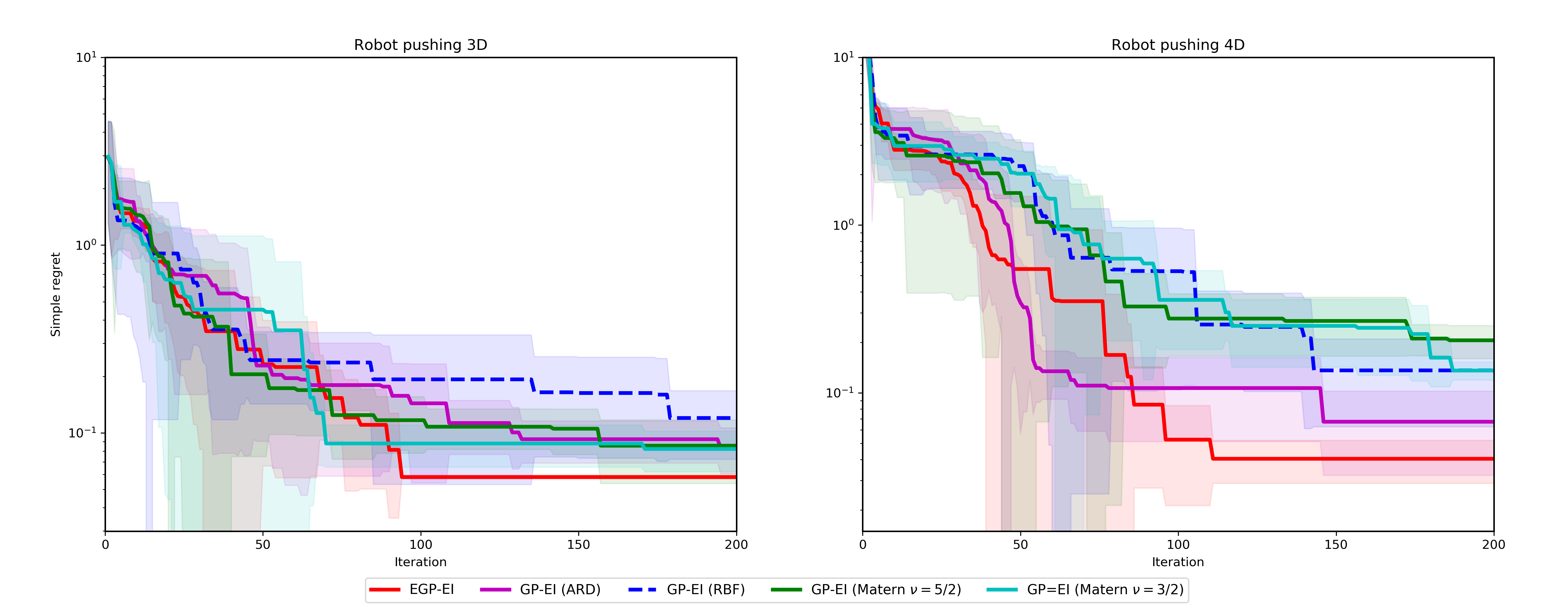}
	\caption{Simple regret of EGP-EI vs. GP-EI with a preselected kernel on \texttt{Robot pushing 3D} and \texttt{Robot pushing 3D} tasks (from left to right). Dictionary has 4 kernels with distinct forms: RBF with(out) ARD and Mat{\'e}rn with $\nu=3/2, 5/2$. }\label{fig:EGP_EI_robot}
\end{figure*}

\begin{table}
	\caption{Feasible values of the hyperparameters for different classification models.} \label{table:stat_NN}
	\begin{center}
		\begin{tabular}{c|c|c}
			\hline
			\hline
	Model &		Hyperparameter & 	Range  \\
			\hline
			\hline 
   \multirow{4}{*}{FNN} &
			No. of neurons at Layer 1 	 & [2,100]	  \\
		&	No. of neurons at Layer 2 	&  [2,100]	   \\
&			Learning rate	&  [$10^{-6}, 10^{-1}$] \\
&			Batch size	   	&  [$2^2, 2^6$]\\
			\hline
   \multirow{2}{*}{SVM} & $C$  & [0.1, 100]\\
    &  $\gamma$ & [0.0001, 10] \\
    \hline
       \multirow{3}{*}{GB} &  Learning rate & [0.1, 10]\\
    &  Subsample ratio & [0.1, 0.99] \\
    &  Max. features ratio & [0.1, 0.99] \\
    \hline
		\end{tabular}
	\end{center}
\end{table}

\subsection{Hyperparameter tuning tasks}
The last test deals with hyperparameter tuning tasks for three classification models, including 
a $2$-layer FNN with ReLU activation function, support vector machine (SVM), and gradient boosting (GB). Note that although the FNN architecture does not yield the state-of-the-art classification performance, it suffices to be used to evaluate different BO methods.
We tested all the competing baselines on \texttt{ Breast cancer}~\cite{BreastCancer}, \texttt{ Iris}~\cite{Iris},  \texttt{Transportation}~\cite{Transportation}, as well as \texttt{Wine}~\cite{Wine} datasets. 
For all the datasets,  $70\%$ of the data are used as the training set, and the remaining are used as the validation set based on which the classification accuracy is calculated. The hyperparameters of the FNN consist of the number of neurons per layer, the learning rate, and the batch size. As for SVM, the values of $C$ and $\gamma$ are to be tuned. For GB, the hyperparameters include the learning rate, subsample ratio, and the ratio of maximum features. Table~\ref{table:stat_NN} summarizes the feasible values of the hyperparameters of the three methods. 
For each set of hyperparameters, the evaluated validation accuracy is obtained as the average of $10$ independent runs on a given dataset.  In FNN training, the number of epochs is chosen to be $20$ and the optimizer used is Adam.

Figs.4-9 depict the best validation accuracy (so far) of the competing methods 
versus the number of iterations for these three classification models. Apparently, EGP-TS with different kernel types or RBF kernels of different lengthscales is shown to outperform the competitors in most of the cases, demonstrating the robustness of the EGP model across tasks.  


\subsection{Preliminary results for EGP-EI} Here, we couple the EGP surrogate with the EI acquisition rule~\cite{jones1998efficient}, yielding the novel EGP-EI approach. In line with the proposed EGP-TS, one could first select a GP model by random sampling based on the weights ${\bf w}_t$, and then implement the EI acquisition function based on the chosen GP model. To benchmark the performance of this advocated EGP-EI, comparison has been made relative to GP-EI with a preselected kernel function. We use BoTorch to implement both EGP-EI and GP-EI with kernel hyperparameters updated every iteration and without RF approximation. As shown in Figs.~10-11, EGP-EI outperforms GP-EI in three out of the four synthetic functions and both of the robotic tasks -- what demonstrates the benefits accompanied with the more expressive EGP model for the EI acquisition function. Rather than sampling a single GP from the EGP, future work includes investigation of the EI rule based on the GP mixture function model (cf. Remark~1).

\section{Conclusions}
This work introduced a non-Gaussian EGP prior with adaptive kernel selection for the sought black-box function in BO. Capitalizing on the RF approximation per GP, acquisition of the subsequent query point is effected via TS, which bypasses the need for design parameters and can readily afford  parallel implementation. Convergence of the proposed EGP-TS algorithm has been established by sublinear cumulative Bayesian regret in both the sequential and parallel settings. Numerical tests demonstrated the merits of EGP-TS relative to existing alternatives. Future work includes investigation of other acquisition functions based on the novel EGP surrogate model, as well as analysis via the notion of frequentist regret.

\section{Proofs}
\vspace*{-0.2cm}
\subsection{Proof of Theorem 1}
Before performing the Bayesian regret analysis for EGP-TS, the following lemma will be first presented. 

\noindent \textbf{Lemma 3.} (Supremum of a GP sample path~\cite{adler1990introduction}). {\it If $f\sim {\cal GP}(0,\kappa^m) $ is a continuous sample path for any $m\in \mathcal{M}$, then $\mathbb{E}[\| f\|_{\infty}] = B < \infty$, and further $\mathbb{E}[\underset{\mathbf{x}\in\mathcal{X}}{\max} |f(\mathbf{x}_*) - f(\mathbf{x})|]\leq 2B$. }

Lemma~ 3 holds when kernels are twice differentiable - what is readily satisfied under Assumption 1.

To bound the cumulative Bayesian regret of EGP-TS, we will rely on its link with the corresponding upper confidence bound algorithm in~\cite{russo2014learning}. Conditioned on GP model $m$ and past data $\mathcal{D}_{t-1}$, the high probability upper confidence bound for $f(\mathbf{x})$ is given by $U^m_t (\mathbf{x}):=\mu_{t-1}^m (\mathbf{x})+\beta_{t}^{1/2} \sigma_{t-1}^m (\mathbf{x})$, where
$\beta_t := 2\log (t^2|\mathcal{X}_t|)$. Here, $\mathcal{X}_t$ is obtained by discretizing each dimension of $\mathcal{X}$ using $n_t = t^2dab\sqrt{\pi}$ equally spaced grids. Thus, $|\mathcal{X}_t| = (n_t)^d$, and
\begin{align}
	\beta_t = 4(d+1)\log t + 2d\log (dab\sqrt{\pi})\approx d\log t  \label{eq:beta_t} \;.
\end{align}
With $[\mathbf{x}]_t$ representing the closest point to $\mathbf{x}$ in $\mathcal{X}_t$, it can be easily verified that
\begin{align}
	\|\mathbf{x} - [\mathbf{x}]_t \|_1 \leq d/n_t, \ \ \forall \mathbf{x}\in \mathcal{X}. \label{eq:x_grid}
\end{align}
Consider next the following decomposition
\begin{align}
	&\mathcal{BR}(T):= \sum_{t=1}^T \mathbb{E}[f(\mathbf{x}_*)-f(\mathbf{x}_t)]\nonumber\\
	& \overset{(a)}{=} \underbrace{\sum_{t=1}^T \mathbb{E}\left[f(\mathbf{x}_*)-f([\mathbf{x}_*]_t)\right]}_{:=A_1} + \underbrace{\sum_{t=1}^T \mathbb{E}\left[f([\mathbf{x}_*]_t-U_t^{m_*}([\mathbf{x}_*]_t)\right]}_{:=A_2} \nonumber\\
	&+ \!\!\underbrace{\sum_{t=1}^T \mathbb{E}\!\left[U_t^{m_*}\!([\mathbf{x}_*]_t)\!-\! U_t^{m_t}([\mathbf{x}_t]_t)\right]}_{:=A_3}  \!+\!\! \underbrace{\sum_{t=1}^T \mathbb{E}\left[U_t^{m_t}([\mathbf{x}_t]_t)\!\!-\!\!f([\mathbf{x}_t]_t)\right]}_{:=A_4}\nonumber\\
	& + \!\underbrace{\sum_{t=1}^T \mathbb{E}\left[f([\mathbf{x}_t]_t)\!-\!\!f(\mathbf{x}_t)\right]}_{:=A_5}\;. \label{eq:BR_decomp}
\end{align}
Since $\{\mathbf{x}_t, m_t\}$ and $\{\mathbf{x}_*, m_*\}$ are identically distributed given $\mathcal{D}_{t-1}$, the fact that $U_t^m (\mathbf{x})$ is a {\it deterministic} function of $\mathcal{D}_{t-1}$ yields $A_3=0$~\cite{russo2014learning, hong2021thompson}.

Next, we will provide an upper bound for $A_1$ and $A_5$ following the proof  in~\cite{kandasamy2018parallelised}. Letting $L_{\max} = \underset{j=\{1,\ldots, d\}}{\sup}\ \ 
\underset{\mathbf{x}\in\mathcal{X}}{\sup}
\left|\frac{\partial f(\mathbf{x})}{\partial x_j} \right|$, the union bound under Assumption~1 implies that
\begin{align}
	{\rm Pr} (L_{\max}\geq c) \leq d a e^{-(c/b)^2} \nonumber
\end{align}
which allows us to obtain
\begin{align}
	& \mathbb{E}[|f(\mathbf{x})-f([\mathbf{x}]_t) |] \leq \mathbb{E}[L\|\mathbf{x}-[\mathbf{x}]_t \|_1] \overset{(a)}{\leq}  \frac{d}{n_t}\mathbb{E}[L_{\rm max}] \nonumber\\
 &\overset{(b)}{=} \frac{d}{n_t}\int_{c=0}^{\infty}{\rm Pr}(L_{\rm max}\geq c)dc \leq \frac{d}{n_t}\int_{c=0}^{\infty} d a e^{-(c/b)^2} dc  \nonumber\\
 &= \frac{\sqrt{\pi}d^2 ab}{2n_t} = \frac{d}{2t^2} \nonumber
\end{align}
where $(a)$ results from~\eqref{eq:x_grid}, and $(b)$ utilizes for $L_{\rm max}\geq 0$ the equality  $\mathbb{E}[L_{\rm max}] = \int_{c=0}^{\infty}{\rm Pr}(L_{\rm max}\geq c)dc $. Hence, $A_1$ and $A_5$ are bounded by
\begin{align}
	A_1=A_5 \leq \sum_{t=1}^T  \frac{d}{2t^2} \leq \frac{\pi^2 d}{12} \;. \label{eq:A_1_5}
\end{align}
Further, $A_2$ can be upper bounded as
\begin{align}
	A_2 &\leq \sum_{t=1}^T\mathbb{E}\left[ \mathbb{I}(f([\mathbf{x}_*]_t) > U_t^{m_*}([\mathbf{x}_*]_t))\left[f([\mathbf{x}_*]_t-U_t^{m_*}([\mathbf{x}_*]_t)\right]  \right] \nonumber\\
	& \leq \sum_{t=1}^T \sum_{m\in\mathcal{M}}\sum_{\mathbf{x}\in\mathcal{X}_t} \mathbb{E}\left[ \mathbb{I}(f(\mathbf{x}) > U_t^{m}(\mathbf{x}))\left[f(\mathbf{x})-U_t^{m}(\mathbf{x})\right]  \right] \nonumber\\
 &\overset{(a)}{=} \sum_{t=1}^T\sum_{m\in\mathcal{M}} \sum_{\mathbf{x}\in\mathcal{X}_t}\frac{\sigma_{t-1}^m (\mathbf{x})}{\sqrt{2\pi}t^2 |\mathcal{X}_t|} \nonumber\\
	&\overset{(b)}{\leq} \sum_{t=1}^T\sum_{m\in\mathcal{M}} \sum_{\mathbf{x}\in\mathcal{X}_t}\frac{1}{\sqrt{2\pi}t^2 |\mathcal{X}_t|} = \frac{\sqrt{2\pi}M}{12} \label{eq:A_2}
\end{align}
where, since $f(\mathbf{x})-U_t^{m} (\mathbf{x})|\mathcal{D}_{t-1} \sim \mathcal{N}\left(-\beta_t^{1/2}\sigma_{t-1}^{m} (\mathbf{x}),(\sigma_{t-1}^{m} (\mathbf{x}))^2\right)$, $(a)$ holds using the identity $\mathbb{E}[r \mathbb{I}(r>0)] = \frac{\sigma }{\sqrt{2\pi}} \exp (-\frac{\mu^2}{2\sigma^2})$ if $r\sim \mathcal{N}(\mu, \sigma^2)$ and $\mu<0$. Inequality $(b)$ is simply due to $\sigma_{t-1}^m (\mathbf{x})\leq 1$.

The last step is to upper bound $A_4$, by constructing a confidence set $\mathcal{C}_t$ for the latent state per slot $t$ so that $m_*\in\mathcal{C}_t$ holds with high probability~\cite{hong2021thompson}.
We will replace $[\mathbf{x}_t]_t$ by $\mathbf{x}_t$ for notational brevity, given that the following result holds for both cases. Consider $\mathcal{C}_t:=\{m\in\mathcal{M}: G_t^m \leq 2\sigma_n \sqrt{N_{t-1}^m \log T}\}$, where  $N_{t-1}^m = \sum_{\tau=1}^{t-1} \mathbb{I}(m_\tau=m)$, and 
\begin{align}
	G_t^m:= \sum_{\tau=1}^{t-1} \mathbb{I}(m_\tau= m) \left(L_\tau^m (\mathbf{x}_\tau)-y_\tau \right)    \;.
\end{align}
Here, $L_t^m (\mathbf{x}) = \mu_{t-1}^m (\mathbf{x}) - \eta \sigma^m_{t-1} (\mathbf{x})$ with $\eta = 2\sqrt{\log T}$ is a lower confidence bound for $f(\mathbf{x})$ conditioned on model $m$.
For later use, we will first present the following two lemmas, whose proofs are deferred to Secs.~7.1.1 and 7.1.2.\\

\noindent \textbf{Lemma 4.} It holds that ${\rm Pr} (m_*\notin \mathcal{C}_t |\mathcal{D}_{t-1} ) \leq 2MT^{-1}, \forall t\in\mathcal{T}:=\{1,\ldots, T\}$.\\

\noindent \textbf{Lemma 5.} It holds that  $\mathbb{E}\left[ \mu_{t-1}^{m_t}(\mathbf{x}_t)-f(\mathbf{x}_t)\right]<3B, \forall m_t\in\mathcal{M}, \mathbf{x}_t\in\mathcal{X}, t\in{\cal T}.$\\

\noindent The following decomposition will be applied towards bounding $A_4$
\begin{align}
	A_4 &= \mathbb{E}\!\left[\sum_{t=1}^T \!\left(U_t^{m_t}(\mathbf{x}_t)\!-\!\mu_{t-1}^{m_t} (\mathbf{x}_t)\right)  \!\right]\!\! +\!  \mathbb{E}\!\left[\sum_{t=1}^T \left(\mu_{t-1}^{m_t}(\mathbf{x}_t)\!-\! f(\mathbf{x}_t)\right)\!\right] \nonumber\\
	&= \underbrace{\sum_{t=1}^T \mathbb{E}[\beta_t^{1/2}\sigma^{m_t}_{t-1} (\mathbf{x}_t)]}_{:=A_{4,1}} \nonumber\\
    &+ \underbrace{ \sum_{t=1}^T {\rm Pr}(m_t\notin {\cal C}_t|{\cal D}_{t-1}) \mathbb{E}\!\left[ \left(\mu_{t-1}^{m_t}(\mathbf{x}_t)\!-\! f(\mathbf{x}_t)\right)| m_t\not\in \mathcal{C}_t\!\right]}_{:=A_{4,3}} \nonumber\\
 &\ \quad  + \underbrace{\sum_{t=1}^T {\rm Pr}(m_t\in {\cal C}_t|{\cal D}_{t-1})\mathbb{E}\left[\left(\mu_{t-1}^{m_t}(\mathbf{x}_t)-f(\mathbf{x}_t)\right) |m_t\in \mathcal{C}_t) \right]}_{:=A_{4,2}}  \label{eq:A_4}
\end{align}
where the last equality is based on the tower property (aka. Law of Total Expectation).

As $m_*$ and $m_t$ are identically distributed given $\mathcal{D}_{t-1}$~\cite{russo2014learning, hong2021thompson}, it is evident from Lemma~4 that ${\rm Pr}(m_t\notin {\cal C}_t|{\cal D}_{t-1})\leq 2MT^{-1}$. Further leveraging Lemma 5, $A_{4,3}$ can be upper bounded by
\begin{align}
	A_{4,3} &= \sum_{t=1}^T  {\rm Pr}(m_t\notin {\cal C}_t|{\cal D}_{t-1}) \mathbb{E}[\left(\mu_{t-1}^{m_t}(\mathbf{x}_t)\!-\! f(\mathbf{x}_t)\right) | (m_t\not\in \mathcal{C}_t)]\nonumber\\
    & \leq 6MB \;.
\end{align}
Meanwhile, we have that
\begin{align}
   &A_{4,2}: \leq \sum_{t=1}^T \mathbb{E}[ \mu^{m_t}_{t-1}({\bf x}_t) - f({\bf x}_t)| m_t\in {\cal C}_t ] \nonumber\\
   & = \sum_{t=1}^T\large( \mathbb{E}[ \left(\mu^{m_t}_{t-1}({\bf x}_t) -L^{m_t}_{t-1}({\bf x}_t)\right)| m_t\in {\cal C}_t ] 
   \nonumber\\
   &+\mathbb{E}[ \left(L^{m_t}_{t-1}({\bf x}_t) -y_t\right)| m_t\in {\cal C}_t ] + \mathbb{E}[\left(y_t- f({\bf x}_t)\right)| m_t\in {\cal C}_t ]\large)\nonumber\\
   & = \sum_{t=1}^T \mathbb{E}[\eta \sigma_{t-1}^{m_t} ({\bf x}_t)]\! + \!\!\!\sum_{m=1}^M  (\mathbb{E}[ (L^{m}_{t_{\rm max}^m\!\!-1}({\bf x}_{t_{\rm max}^m}\!) \!-\!y_{t_{\rm max}^m}\!) | m\!\in\! {\cal C}_{t_{\rm max}^m} ] \!\!\!\!\!\!\!\nonumber\\
   &\quad + \sum_{t \in {\cal T}^m_T\setminus t_{\rm max}^m }\!\!\!\!\!\!\mathbb{E}[ \left(L^{m}_{t-1}({\bf x}_t) -y_t\right) | m\in {\cal C}_t ])\nonumber\\
   & \overset{(a)}{\leq} \sum_{t=1}^T \mathbb{E}[\eta \sigma_{t-1}^{m_t} ({\bf x}_t)]  + 3MB + \sum_{m=1}^M \mathbb{E}[G_{t_{\rm max}^m}^m] \nonumber\\
   &\overset{(b)}{\leq} \sum_{t=1}^T \mathbb{E}[\eta \sigma_{t-1}^{m_t} ({\bf x}_t)]  \!+\! 3MB \!+\! \sum_{m=1}^M \!\!\mathbb{E}\left[2\sigma_n\sqrt{(T_m\!-\!1)\log T}\right] \nonumber\\
     &\overset{(c)}{\leq} \sum_{t=1}^T \mathbb{E}[\eta \sigma_{t-1}^{m_t} ({\bf x}_t)]  + 3MB + 2\sigma_n\sqrt{MT\log T}
\end{align}
where ${\cal T}_T^m:=\{t\in {\cal T}:m_t=m \}$, $T_m:=|\mathcal{T}_T^m|$, and $t_{\rm max}^m$ is the maximum slot index that $m$ is selected, namely $t_{\rm max}^m:=\max_{t\in {\cal T}_T^m} t$.
The inequality $(a)$ holds since $\mathbb{E}[ (L^{m}_{t_{\rm max}^m-1}({\bf x}_{t_{\rm max}^m}) -y_{t_{\rm max}^m}) | m\in {\cal C}_{t_{\rm max}^m} ]\leq 3B$ for any $t_{\rm max}^m$ given Lemma 5. We want to clarify that
although $t_{\rm max}^m$ is a random variable, the inequality in (a) holds for any value of $t_{\rm max}^m$. Inequality $(b)$ is from the definition of ${\cal C}_t$ and $(c)$ leverages the Cauchy-Schwarz inequality to yield
\begin{align}
	\sum_{m=1}^M\sqrt{T_m}\leq \sqrt{M\sum_{m=1}^M T_m} = \sqrt{MT}\;. \end{align}
Putting together the bounds for $A_{1}-A_{5}$, the cumulative Bayesian regret of EGP-TS over $T$ evaluations is bounded by
\begin{align}
	{\cal BR}(T) &\leq  (\eta+\beta_T^{1/2})\sum_{t=1}^T \mathbb{E}[\sigma^{m_t}_{t-1} (\mathbf{x}_t)] + 2\sigma_n \sqrt{MT\log T} \nonumber\\
 & \ \quad + 6MB + \frac{\pi^2 d}{6} + \frac{\sqrt{2\pi}M}{12}  \label{eq:BR_proof}
\end{align}
where the first term can be bounded as
\begin{align}
	 &\sum_{t=1}^T \mathbb{E}[\sigma^{m_t}_{t-1} (\mathbf{x}_t)] 
	\leq \sum_{m=1}^{M} \mathbb{E}\left[\sum_{t\in\mathcal{T}_T^m}  \sigma^{m}_{t-1} (\mathbf{x}_t) \right]   \nonumber\\
	 &\overset{(a)}{\leq}\sum_{m=1}^{M} \mathbb{E}\left[\sum_{t=1}^{T_m} \sigma^{m}_{t-1} (\mathbf{x}_t) \right]  \overset{(b)}{\leq}\sum_{m=1}^M\mathbb{E} \left(T_m \sum_{t=1}^{T_m}  \left(\sigma_{t-1}^m (\mathbf{x}_t)\right)^2\right)^{1/2} \nonumber\\
	& \overset{(c)}{\leq}  \sum_{m=1}^M \left(\frac{2{T_m}  \gamma_{T_m} }{\log\left(1+\sigma_n^{-2}\right) } \right)^{1/2}  \overset{(d)}{\leq}  \sum_{m=1}^M \left(\frac{2T_m^{1+c}}{\log\left(1+\sigma_n^{-2}\right) } \right)^{1/2} \nonumber\\
	& \overset{(e)}{\leq}  \left(\frac{2M T^{1+c}}{\log\left(1+\sigma_n^{-2}\right)} \right)^{1/2} \label{eq:A4_MIG} 
\end{align}
where $(a)$ holds since $\sigma_t^m (\mathbf{x})$ decreases as $t$ grows; $(b)$ is due to the Cauchy-Schwarz inequality; $(c)$ leverages Lemmas 5.3 and 5.4 of~\cite{srinivas2012information} that bound the sum of posterior variances via the MIG; $(d)$ follows upon bounding $\gamma_{T_m}$ using Lemma~1; and, $(e)$ holds upon utilizing the following inequality based on Cauchy-Schwarz inequality
\begin{align}
	 \sum_{m=1}^M \left(T_m^{1+c}\right)^{1/2} &\leq \left(M\sum_{m=1}^M T_m^{1+c} \right)^{1/2} \nonumber\\
  &\leq \left(M\left(\sum_{m=1}^M T_m\right)^{1+c} \right)^{1/2} = \left(M T^{1+c} \right)^{1/2} \nonumber.
\end{align}
Upon plugging in~\eqref{eq:A4_MIG} into \eqref{eq:BR_proof}, Theorem~1 holds with $\eta = 2\sqrt{\log T}$ and $\beta_T^{1/2}\approx \sqrt{d\log T}$.

\subsubsection{Proof for Lemma 4}
For $m_*\in\mathcal{M}$, define the following event at slot $t$ 
\begin{align}
	\mathcal{E}_t^{m_*}:= \{|f(\mathbf{x})-\mu^{m_*}_{t-1}(\mathbf{x})|\leq \eta \sigma_{t-1}^{m_*} (\mathbf{x}) \}
\end{align}
the collection of which over $T$ slots is $\mathcal{E}^{m_*}_{1:T}:=\cap_{t=1}^T \mathcal{E}_t^{m_*} $. With $\bar{\mathcal{E}}^{m_*}_{1:T}$ representing its complement, it follows that
\begin{align}
	 &\mathbb{E} [I(\bar{\mathcal{E}}^{m_*}_{1:T})]   \leq \sum_{t=1}^T \sum_{m\in\mathcal{M}} \mathbb{E}\left[ \mathbb{E}_{t-1}[ \mathbb{I}(\bar{\mathcal{E}}^{m}_t)] \right] \nonumber\\
 &= \sum_{t=1}^T \sum_{m\in\mathcal{M}} \mathbb{E}\left[ {\rm Pr}_{t-1} \left(|\mu^{m}_{t-1}(\mathbf{x})-f(\mathbf{x})|> \eta \sigma_{t-1}^m (\mathbf{x})\right) \right]  \nonumber\\
	&\overset{(a)}{\leq} MT^{-1}
\end{align}
where $(a)$ comes from the inequality ${\rm Pr} (|r|>\eta) \leq e^{-\eta^2/2}$ with $r =|\mu^{m}_{t-1}(\mathbf{x})-f(\mathbf{x})| /\sigma_{t-1}^m (\mathbf{x})\sim \mathcal{N}(0,1)$ and $\eta = 2\sqrt{\log T}$.

Since $n_\tau = f(\mathbf{x}_\tau) -y_\tau\sim \mathcal{N}(0,\sigma_n^2)$, $\{n_\tau\}_{\tau\in\mathcal{T}_{t}^m}$ is then a martingale difference sequence w.r.t. $\{\mathcal{D}_\tau\}_{\tau \in \mathcal{T}_{t}^m}$, where $\mathcal{T}_{t}^m:=\{\tau|m_\tau=m, \tau\in \{1,\ldots,t\}\}$. 
\begin{align}
	&G_t^m I(\mathcal{E}^m_{1:T}) \\
 &= \sum_{\tau\in \mathcal{T}_{t}^m} (L_\tau^m(\mathbf{x}_\tau)-y_\tau) \mathbb{I}(|f(\mathbf{x}_\tau)-\mu^{m}_{\tau-1}(\mathbf{x}_\tau)|\leq \eta \sigma_{\tau-1}^{m} (\mathbf{x}_\tau))\nonumber\\
	&= \sum_{\tau\in \mathcal{T}_{t}^m} (L_\tau^m(\mathbf{x}_\tau)-y_\tau) \mathbb{I}(L_\tau^m(\mathbf{x}_\tau)<f(\mathbf{x}_\tau))  \leq \sum_{\tau\in \mathcal{T}_{t}^m} n_\tau  \;.\nonumber
\end{align}
For any $m\in\mathcal{M}$ and $t\in\mathcal{T}$,  $u=|\mathcal{T}_t^m|$ is random and takes value from $\{1,\ldots,t-1 \}$. For any $u$, 
Azuma's inequality yields
\begin{align}
	 & {\rm Pr}_{t-1} (G_t^m I(\mathcal{E}^m_{1:T})\geq 2\sigma_n \sqrt{u \log T})\nonumber\\
  & \leq {\rm Pr} \left(\sum_{\tau\in \mathcal{T}_{t}^m} n_\tau \geq 2\sigma_n\sqrt{u\log T} \right)\leq \exp(-2\log T) = T^{-2}\nonumber
\end{align}
based on which, we arrive at 
\begin{align}
&	{\rm Pr}_t (m_*\not\in \mathcal{C}_t)  \leq \sum_{m\in \mathcal{M}}\sum_{u=1}^{t-1} {\rm Pr}(G_t^m\geq 2\sigma_n\sqrt{u\log T}) \nonumber\\
	& \leq\!\! \sum_{m\in\mathcal{M}} \sum_{u=1}^{t-1} \!\mathbb{E}\!\left[{\rm Pr}_{t-1} \!\left(G_t^m \mathbb{I}(\mathcal{E}^m_{1:T})\!\geq\! 2 \sigma_n\sqrt{u\log T} \right) \right]
	\!+\! {\rm Pr}(\bar{\mathcal{E}}^{m_*}_{1:T})\nonumber\\
 &\leq 2MT^{-1} \;.
\end{align}
thus finalizing the proof of Lemma 4.

\subsubsection{ Proof for Lemma 5}
Subtracting and adding $\mathbb{E}_{t-1}[\mu_{t-1}^{m_*}(\mathbf{x}_*)]$ and $f(\mathbf{x}_*)$ yields the following decomposition
\begin{align}
 & \mathbb{E}[\mu^{m_t}_{t-1}(\mathbf{x}_t)-f(\mathbf{x}_t)]
  =\mathbb{E}[\mu^{m_t}_{t-1}(\mathbf{x}_t)- \mu^{m_*}_{t-1}(\mathbf{x}_*)]\nonumber\\
  &\quad \quad+{\mathbb{E}[\mu^{m_*}_{t-1}(\mathbf{x}_*)
    -f(\mathbf{x}_*)]}
  +{\mathbb{E}[f(\mathbf{x}_*)-f(\mathbf{x}_t)]}. \nonumber
\end{align}
Since $\{\mathbf{x}_t, m_t\}$ and $\{\mathbf{x}_*, m_*\}$ are identically distributed conditioned on $\mathcal{D}_{t-1}$, it holds that 
\begin{align}
	\mathbb{E}[\mu_{t-1}^{m_t}(\mathbf{x}_t)] &= \mathbb{E}[\mathbb{E}_{t-1}[\mu_{t-1}^{m_t}(\mathbf{x}_t)]]=\mathbb{E}[\mathbb{E}_{t-1}[\mu_{t-1}^{m_*}(\mathbf{x}_*)]] \nonumber\\
    &= \mathbb{E}[\mu_{t-1}^{m_*}(\mathbf{x}_*)]\;. \nonumber
\end{align}
Further, $(f(\mathbf{x})-\mu^{m_*}_{t-1}(\mathbf{x}))| {\cal D}_{t-1}$ is a centered posterior GP, namely, $(f(\mathbf{x})-\mu^{m_*}_{t-1}(\mathbf{x}))| {\cal D}_{t-1}
\sim {\cal GP}(0,\kappa^{m_*}_{t-1})$, where
$\kappa^{m_*}_{t-1}(\mathbf{x},\mathbf{x})\le\kappa^{m_*}(\mathbf{x},\mathbf{x})$.  Thanks to Lemma 3, the following inequality holds 
\begin{align}
  \mathbb{E}[\mu^{m_*}_{t-1}(\mathbf{x})
    -f(\mathbf{x})] & = \mathbb{E}[\mathbb{E}_{t-1}[  \mu^{m_*}_{t-1}(\mathbf{x})
    -f(\mathbf{x})]]\nonumber\\
    &\leq \mathbb{E}[\mathbb{E}_{t-1}[|f(\mathbf{x})-\mu^{m_*}_{t-1}(\mathbf{x})|]]
\leq B \nonumber
\end{align}
for any point ${\bf x}$
in ${\cal X}$, whether
fixed or random, including ${\bf x}^{*}$. That is, $
     \mathbb{E}[\mu^{m_*}_{t-1}(\mathbf{x}_*)
    -f(\mathbf{x}_*)] \leq B$.

Leveraging Lemma 3, one then arrives at
 $\mathbb{E}[f(\mathbf{x}_*) -f(\mathbf{x}_t)] \leq 2B$,
which yields
$ \mathbb{E}\ [\mu^{m_t}_{t-1}(\mathbf{x}_t)-f(\mathbf{x}_t)]\;\le\;B+2B = 3B$.   

\subsection{Proof of Theorem 2}
For the asynchronous parallel setting, the upper confidence bound for the $t$th function evaluation is given by
\begin{align}
	\bar{U}_t^m (\mathbf{x}):= \mu_{\mathcal{D}_{t-1}}^m (\mathbf{x}) +\beta_t^{1/2} \sigma_{\mathcal{D}_{t-1}}^m (\mathbf{x}) \label{eq:UCB_asy}
\end{align}
where $\mathcal{D}_{t-1}$ contains all the acquired data before evaluation index $t$ is assigned. Here, $|\mathcal{D}_{t-1}|=t-K$ for $t>K$, and $|\mathcal{D}_{t-1}|=0$ for $t\leq K$.

Leveraging a decomposition similar to that in~\eqref{eq:BR_decomp}, $A_1$--$A_3$ and $A_5$ could be derived as in~Sec.~A. Upon replacing the subscript $t-1$ of $\mu$ and $\sigma$ by $\mathcal{D}_{t-1}$, the term $A_4$ can be bounded as in~\eqref{eq:A_4}, that is 
\begin{align}
	A_4 
	&\leq \sum_{t=1}^T \mathbb{E}[(\beta_t^{1/2}+\eta)\sigma^{m_t}_{\mathcal{D}_{t-1}} (\mathbf{x}_t)] + 2\sigma_n \sqrt{MT\log T}+ 9M\!B
\end{align}
where the first term can be further bounded based on Lemma~2 and \eqref{eq:A4_MIG} as
\begin{align}
	& \sum_{t=1}^T \mathbb{E}[(\beta_t^{1/2}+\eta)\sigma^{m_t}_{\mathcal{D}_{t-1}} (\mathbf{x}_t)] 
	\overset{(a)}{\leq} (\beta_T^{1/2}+\eta) \sum_{t=1}^T \mathbb{E}[\rho_K^{1/2}\sigma^{m_t}_{t-1}] \nonumber\\
 &\leq (2+\sqrt{d} ) \left(\frac{2\rho_K M T^{c+1}\log T}{\log\left(1+\sigma_n^{-2}\right)} \right)^{1/2}\;.
\end{align}
Thus, the cumulative Bayesian regret for parallel EGP-TS in the asynchronous setup can be established as in Theorem~2.

\subsection{Proof of Theorem 3}
The proof of Theorem~3 entails introducing
\begin{align}
	V_t^m (\mathbf{x}):=\mu_{\mathcal{D}_{t-1}}^m (\mathbf{x}) + \beta_{t+K-1}^{1/2} \sigma_{t-1}^m (\mathbf{x})
\end{align}
based on which the cumulative Bayesian regret can be decomposed after using~\eqref{eq:UCB_asy} as  (cf. \eqref{eq:BR_decomp})
\vspace*{-0.2cm}
\begin{align}
	&\mathcal{BR}^{\rm syn}(T):= \sum_{t=1}^T \mathbb{E}[f(\mathbf{x}_*)-f(\mathbf{x}_t)]\label{eq:BR_decomp_syn}\\
	& \overset{(a)}{=} \underbrace{\sum_{t=1}^T \mathbb{E}\left[f(\mathbf{x}_*)-f([\mathbf{x}_*]_t)\right]}_{:=C_1} + \underbrace{\sum_{t=1}^T \mathbb{E}\left[f([\mathbf{x}_*]_t-\bar{U}_t^{m_*}([\mathbf{x}_*]_t)\right]}_{:=C_2} \nonumber\\
	& + \underbrace{\sum_{t=1}^T \mathbb{E}\left[\bar{U}_t^{m_*}([\mathbf{x}_*]_t)-V_t^{m_*}([\mathbf{x}_*]_t)\right]}_{:=C_3}\nonumber\\
	&+ \underbrace{\sum_{t=1}^T \mathbb{E}\left[V_t^{m_*}([\mathbf{x}_*]_t)- V_t^{m_t}([\mathbf{x}_t]_t)\right]}_{:=C_4}\nonumber\\
	& + \underbrace{\sum_{t=1}^T \mathbb{E}\left[V_t^{m_t}([\mathbf{x}_t]_t)-f([\mathbf{x}_t]_t)\right]}_{:=C_5} + \underbrace{\sum_{t=1}^T \mathbb{E}\left[f([\mathbf{x}_t]_t)-f(\mathbf{x}_t)\right]}_{:=C_6}\;. \nonumber
\end{align}
As with the proof of Theorem~1, it follows that
\begin{align}
	C_1 = C_6 \leq \frac{\pi^2 d}{12}, \quad C_4 = 0, \quad C_2 \leq \frac{\sqrt{2\pi}M}{12}. 
\end{align}
Next, we will further bound $C_3$ and $C_5$, starting with
\begin{align}
	& C_3 = \sum_{t=1}^{K-1} \beta_t^{1/2} \mathbb{E}[\sigma^{m_*}_{{\cal D}_{t-1}}([\mathbf{x}_*]_t)] -\!\!\! \sum_{T-K+1}^{T}\!\! \beta_{t+M-1}^{1/2} \mathbb{E}[\sigma^{m_*}_{t-1}([\mathbf{x}_*]_t)] \nonumber\\
	&\ + \sum_{t=K}^{T-K} \beta_{t}^{1/2} \mathbb{E} [\sigma^{m_*}_{{\cal D}_{t-1}}([\mathbf{x}_*]_t) - \sigma^{m_*}_{t-K} ([\mathbf{x}_*]_t) ] \overset{(a)}{\leq} (K-1)\beta_{K-1}^{1/2} \nonumber
\end{align}
where $(a)$ holds since $\sigma^{m_*}_{{\cal D}_{t-1}}([\mathbf{x}_*]_t) \leq \sigma^{m_*}_{t-K} ([\mathbf{x}_*]_t) $, and $ 0<\sigma^{m_*}_{t} (\mathbf{x})\leq 1$.

Lastly, $C_5$ can be bounded as
\begin{align}
	   &C_5  =\mathbb{E}\left[\sum_{t=1}^T \left(V_t^{m_t}(\mathbf{x}_t)-\mu_{\mathcal{D}_{t-1}}^{m_t} (\mathbf{x}_t)\right)  \right] \nonumber\\
  &\ \qquad +  \mathbb{E}\left[\sum_{t=1}^T \left(\mu_{\mathcal{D}_{t-1}}^{m_t}(\mathbf{x}_t)-f(\mathbf{x}_t)\right)\right] \nonumber\\
	&\leq \sum_{t=1}^T \mathbb{E}[\beta_{t+K-1}^{1/2}\sigma^{m_t}_{t-1} (\mathbf{x}_t)] \nonumber\\
 &+\sum_{t=1}^T {\rm Pr}(m_t\not\in \mathcal{C}_t|{\cal D}_{t-1})  \mathbb{E} \left[\left(\mu_{\mathcal{D}_{t-1}}^{m_t}(\mathbf{x}_t)-f(\mathbf{x}_t)\right) \big|(m_t\notin \mathcal{C}_t)  \right] \nonumber\\
 & + \sum_{t=1}^T \mathbb{E}\left[\left(\mu_{\mathcal{D}_{t-1}}^{m_t}(\mathbf{x}_t)-f(\mathbf{x}_t)\right) \big|(m_t\in \mathcal{C}_t) \right]\!  \nonumber\\
	& \leq \sum_{t=1}^T \mathbb{E}[\beta_{t+K-1}^{1/2} \sigma^{m_t}_{t-1} (\mathbf{x}_t) + \eta\sigma^{m_t}_{\mathcal{D}_{t-1}} (\mathbf{x}_t) ] \nonumber\\
 & \qquad + 2\sigma_n \sqrt{MT\log T}+ 9MB\nonumber\\
	& \leq \sum_{t=1}^T \mathbb{E}[\beta_{T+K-1}^{1/2} \sigma^{m_t}_{t-1} (\mathbf{x}_t) + \eta\rho_K^{1/2}\sigma^{m_t}_{t-1} (\mathbf{x}_t) ] \nonumber\\
 & \qquad + 2\sigma_n \sqrt{MT\log T}+ 9MB 	\label{eq:C_5}
\end{align}
which, based on the derivation of~\eqref{eq:A4_MIG}, and the bounds of other factors, yields the regret bound in Theorem~3.

\ifCLASSOPTIONcompsoc
  \section*{Acknowledgments}
\else
  \section*{Acknowledgment}
\fi

The authors would like to thank the anonymous reviewers for their constructive feedback. We also gratefully acknowledge the support from NSF grants 1901134, 2128593, 2126052, 2212318, and 2220292.

\ifCLASSOPTIONcaptionsoff
  \newpage
\fi



%

\bibliographystyle{IEEEtran}
\bibliography{EGP_BO}

\begin{thebibliography}{10}
\providecommand{\url}[1]{#1}
\csname url@samestyle\endcsname
\providecommand{\newblock}{\relax}
\providecommand{\bibinfo}[2]{#2}
\providecommand{\BIBentrySTDinterwordspacing}{\spaceskip=0pt\relax}
\providecommand{\BIBentryALTinterwordstretchfactor}{4}
\providecommand{\BIBentryALTinterwordspacing}{\spaceskip=\fontdimen2\font plus
\BIBentryALTinterwordstretchfactor\fontdimen3\font minus
  \fontdimen4\font\relax}
\providecommand{\BIBforeignlanguage}[2]{{%
\expandafter\ifx\csname l@#1\endcsname\relax
\typeout{** WARNING: IEEEtran.bst: No hyphenation pattern has been}%
\typeout{** loaded for the language `#1'. Using the pattern for}%
\typeout{** the default language instead.}%
\else
\language=\csname l@#1\endcsname
\fi
#2}}
\providecommand{\BIBdecl}{\relax}
\BIBdecl

\bibitem{snoek2012practical}
J.~Snoek, H.~Larochelle, and R.~P. Adams, ``Practical {B}ayesian optimization
  of machine learning algorithms,'' \emph{Proc. Adv. Neural Inf. Process.
  Syst.}, vol.~25, 2012.

\bibitem{korovina2020chembo}
K.~Korovina, S.~Xu, K.~Kandasamy, W.~Neiswanger, B.~Poczos, J.~Schneider, and
  E.~Xing, ``Chembo: {B}ayesian optimization of small organic molecules with
  synthesizable recommendations,'' \emph{Proc. Int. Conf. Artif. Intel. and
  Stats.}, pp. 3393--3403, 2020.

\bibitem{cully2015robots}
A.~Cully, J.~Clune, D.~Tarapore, and J.-B. Mouret, ``Robots that can adapt like
  animals,'' \emph{Nature}, vol. 521, no. 7553, pp. 503--507, 2015.

\bibitem{shahriari2015taking}
B.~Shahriari, K.~Swersky, Z.~Wang, R.~P. Adams, and N.~De~Freitas, ``Taking the
  human out of the loop: A review of {B}ayesian optimization,'' \emph{Proc.
  IEEE}, vol. 104, no.~1, pp. 148--175, 2015.

\bibitem{frazier2018tutorial}
P.~I. Frazier, ``A tutorial on {B}ayesian optimization,'' \emph{arXiv preprint
  arXiv:1807.02811}, 2018.

\bibitem{turner2021bayesian}
R.~Turner, D.~Eriksson, M.~McCourt, J.~Kiili, E.~Laaksonen, Z.~Xu, and
  I.~Guyon, ``{B}ayesian optimization is superior to random search for machine
  learning hyperparameter tuning: {A}nalysis of the black-box optimization
  challenge 2020,'' \emph{arXiv preprint arXiv:2104.10201}, 2021.

\bibitem{wang2018batched}
Z.~Wang, C.~Gehring, P.~Kohli, and S.~Jegelka, ``Batched large-scale {B}ayesian
  optimization in high-dimensional spaces,'' \emph{Proc. Int. Conf. Artif.
  Intel. and Stats.}, pp. 745--754, 2018.

\bibitem{feurer2018scalable}
M.~Feurer, B.~Letham, and E.~Bakshy, ``Scalable meta-learning for {B}ayesian
  optimization using ranking-weighted {G}aussian process ensembles,'' in
  \emph{AutoML Workshop at ICML}, vol.~7, 2018.

\bibitem{hoffman2011portfolio}
M.~Hoffman, E.~Brochu, N.~de~Freitas \emph{et~al.}, ``Portfolio allocation for
  bayesian optimization.'' \emph{Proc. Conf. Uncerntainty in Artif. Intel.},
  pp. 327--336, 2011.

\bibitem{shahriari2014entropy}
B.~Shahriari, Z.~Wang, M.~W. Hoffman, A.~Bouchard-C{\^o}t{\'e}, and
  N.~de~Freitas, ``An entropy search portfolio for {B}ayesian optimization,''
  \emph{arXiv preprint arXiv:1406.4625}, 2014.

\bibitem{thompson1933likelihood}
W.~R. Thompson, ``On the likelihood that one unknown probability exceeds
  another in view of the evidence of two samples,'' \emph{Biometrika}, vol.~25,
  no. 3/4, pp. 285--294, 1933.

\bibitem{chapelle2011empirical}
O.~Chapelle and L.~Li, ``An empirical evaluation of {T}hompson sampling,''
  \emph{Proc. Adv. Neural Inf. Process. Syst.}, vol.~24, pp. 2249--2257, 2011.

\bibitem{russo2014learning}
D.~Russo and B.~Van~Roy, ``Learning to optimize via posterior sampling,''
  \emph{Mathematics of Operations Research}, vol.~39, no.~4, pp. 1221--1243,
  2014.

\bibitem{mutny2019efficient}
M.~Mutn{\`y} and A.~Krause, ``Efficient high dimensional {B}ayesian
  optimization with additivity and quadrature {F}ourier features,'' \emph{Proc.
  Adv. Neural Inf. Process. Syst.}, pp. 9005--9016, 2019.

\bibitem{nguyen2020bayesian}
D.~Nguyen, S.~Gupta, S.~Rana, A.~Shilton, and S.~Venkatesh, ``Bayesian
  optimization for categorical and category-specific continuous inputs,''
  \emph{Proc. AAAI Conf. Artif. Intel.}, vol.~34, no.~04, pp. 5256--5263, 2020.

\bibitem{gopakumar2018algorithmic}
S.~Gopakumar, S.~Gupta, S.~Rana, V.~Nguyen, and S.~Venkatesh, ``Algorithmic
  assurance: {A}n active approach to algorithmic testing using {B}ayesian
  optimisation,'' \emph{Proc. Adv. Neural Inf. Process. Syst.}, pp. 5470--5478,
  2018.

\bibitem{kandasamy2018parallelised}
K.~Kandasamy, A.~Krishnamurthy, J.~Schneider, and B.~P{\'o}czos, ``Parallelised
  {B}ayesian optimisation via {T}hompson sampling,'' \emph{Proc. Int. Conf.
  Artif. Intel. and Stats.}, pp. 133--142, 2018.

\bibitem{hernandez2017parallel}
J.~M. Hern{\'a}ndez-Lobato, J.~Requeima, E.~O. Pyzer-Knapp, and
  A.~Aspuru-Guzik, ``Parallel and distributed {T}hompson sampling for
  large-scale accelerated exploration of chemical space,'' \emph{Proc. Int.
  Conf. Mach. Learn.}, pp. 1470--1479, 2017.

\bibitem{chowdhury2017kernelized}
S.~R. Chowdhury and A.~Gopalan, ``On kernelized multi-armed bandits,'' pp.
  844--853, 2017.

\bibitem{vakili2020scalable}
S.~Vakili, V.~Picheny, and A.~Artemev, ``Scalable {T}hompson sampling using
  sparse {G}aussian process models,'' \emph{Proc. Adv. Neural Inf. Process.
  Syst.}, 2021.

\bibitem{hong2021thompson}
J.~Hong, B.~Kveton, M.~Zaheer, M.~Ghavamzadeh, and C.~Boutilier, ``Thompson
  sampling with a mixture prior,'' \emph{arXiv preprint arXiv:2106.05608},
  2021.

\bibitem{desautels2014parallelizing}
T.~Desautels, A.~Krause, and J.~W. Burdick, ``Parallelizing
  exploration-exploitation tradeoffs in {G}aussian process bandit
  optimization,'' \emph{J. Mach. Learn. Res.}, vol.~15, pp. 3873--3923, 2014.

\bibitem{wang2016parallel}
J.~Wang, S.~C. Clark, E.~Liu, and P.~I. Frazier, ``Parallel {B}ayesian global
  optimization of expensive functions,'' \emph{arXiv preprint
  arXiv:1602.05149}, 2016.

\bibitem{teng2020scalable}
T.~Teng, J.~Chen, Y.~Zhang, and B.~K.~H. Low, ``Scalable variational {B}ayesian
  kernel selection for sparse {G}aussian process regression,'' \emph{Proc. AAAI
  Conf. Artif. Intel.}, vol.~34, no.~04, pp. 5997--6004, 2020.

\bibitem{duvenaud2013structure}
D.~Duvenaud, J.~Lloyd, R.~Grosse, J.~Tenenbaum, and G.~Zoubin, ``Structure
  discovery in nonparametric regression through compositional kernel search,''
  \emph{Proc. Int. Conf. Mach. Learn.}, pp. 1166--1174, 2013.

\bibitem{kim2018scaling}
H.~Kim and Y.~W. Teh, ``Scaling up the automatic statistician: Scalable
  structure discovery using {G}aussian processes,'' \emph{Proc. Int. Conf.
  Artif. Intel. and Stats.}, pp. 575--584, 2018.

\bibitem{malkomes2016bayesian}
G.~Malkomes, C.~Schaff, and R.~Garnett, ``Bayesian optimization for automated
  model selection,'' \emph{Proc. Adv. Neural Inf. Process. Syst.}, 2016.

\bibitem{lu2020ensemble}
Q.~Lu, G.~Karanikolas, Y.~Shen, and G.~B. Giannakis, ``Ensemble {G}aussian
  processes with spectral features for online interactive learning with
  scalability,'' \emph{Proc. Int. Conf. Artif. Intel. and Stats.}, pp.
  1910--1920, 2020.

\bibitem{devlin2018bert}
J.~Devlin, M.-W. Chang, K.~Lee, and K.~Toutanova, ``Bert: Pre-training of deep
  bidirectional transformers for language understanding,'' \emph{arXiv preprint
  arXiv:1810.04805}, 2018.

\bibitem{Rasmussen2006gaussian}
C.~E. Rasmussen and C.~K. Williams, \emph{Gaussian processes for machine
  learning}.\hskip 1em plus 0.5em minus 0.4em\relax MIT press Cambridge, MA,
  2006.

\bibitem{rahimi2008random}
A.~Rahimi and B.~Recht, ``Random features for large-scale kernel machines,''
  \emph{Proc. Adv. Neural Inf. Process. Syst.}, pp. 1177--1184, 2008.

\bibitem{wilson2020efficiently}
J.~Wilson, V.~Borovitskiy, A.~Terenin, P.~Mostowsky, and M.~Deisenroth,
  ``Efficiently sampling functions from {G}aussian process posteriors,''
  \emph{Proc. Int. Conf. Mach. Learn.}, pp. 10\,292--10\,302, 2020.

\bibitem{quia2010sparse}
M.~L\'azaro-Gredilla, J.~Qui\~nonero Candela, C.~E. Rasmussen, and
  A.~Figueiras-Vidal, ``Sparse spectrum {G}aussian process regression,''
  \emph{J. Mach. Learn. Res.}, vol.~11, no. Jun, pp. 1865--1881, 2010.

\bibitem{jones1998efficient}
D.~R. Jones, M.~Schonlau, and W.~J. Welch, ``Efficient global optimization of
  expensive black-box functions,'' \emph{Journal of Global optimization},
  vol.~13, no.~4, pp. 455--492, 1998.

\bibitem{srinivas2012information}
N.~Srinivas, A.~Krause, S.~M. Kakade, and M.~W. Seeger, ``Information-theoretic
  regret bounds for {G}aussian process optimization in the bandit setting,''
  \emph{IEEE Trans. Inf. Theory}, vol.~58, no.~5, pp. 3250--3265, 2012.

\bibitem{ghosal2006posterior}
S.~Ghosal and A.~Roy, ``Posterior consistency of {G}aussian process prior for
  nonparametric binary regression,'' \emph{The Annals of Statistics}, vol.~34,
  no.~5, pp. 2413--2429, 2006.

\bibitem{wang2017max}
Z.~Wang and S.~Jegelka, ``Max-value entropy search for efficient {B}ayesian
  optimization,'' \emph{Proc. Int. Conf. Mach. Learn.}, pp. 3627--3635, 2017.

\bibitem{BreastCancer}
``Breast {Cancer} dataset,''
  \url{https://archive.ics.uci.edu/ml/datasets/breast+cancer}.

\bibitem{Iris}
``Iris dataset,'' \url{https://archive.ics.uci.edu/ml/datasets/iris}.

\bibitem{Transportation}
``Transportation dataset,'' \url{https://gisdata.mn.gov}.

\bibitem{Wine}
``Wine dataset,'' \url{https://archive.ics.uci.edu/ml/datasets/wine}.

\bibitem{adler1990introduction}
R.~J. Adler, ``An introduction to continuity, extrema, and related topics for
  general {G}aussian processes.''\hskip 1em plus 0.5em minus 0.4em\relax IMS,
  1990.

\bibitem{rudin1964principles}
W.~Rudin, \emph{Principles of {M}athematical {A}nalysis}.\hskip 1em plus 0.5em
  minus 0.4em\relax McGraw-hill New York, 1964, vol.~3.

\end{thebibliography}

\begin{IEEEbiography}[{\includegraphics[width=1in,height=1.25in]{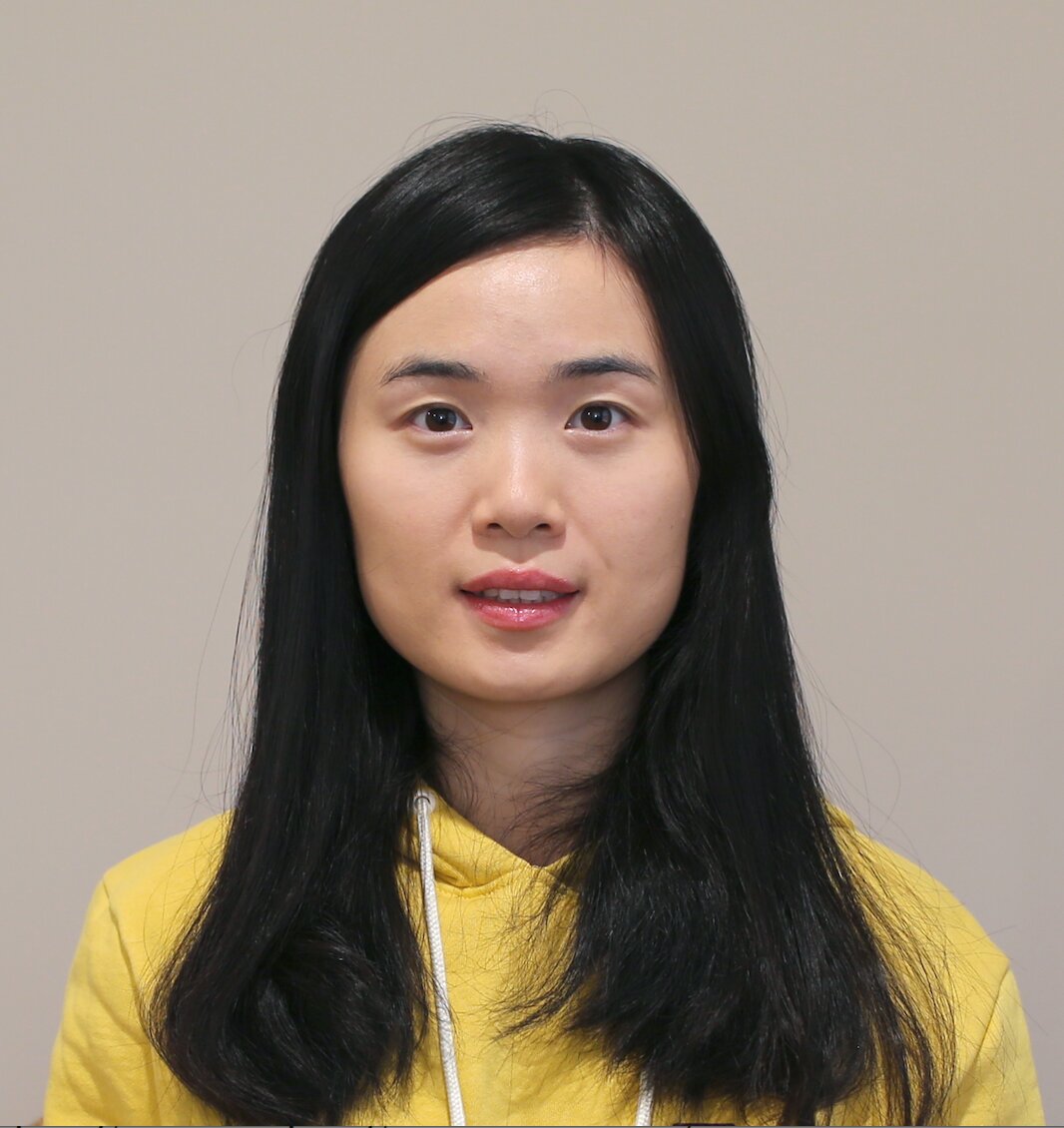}}]{Qin Lu (M'18)}
received her Ph.D. degree in electrical engineering from University of Connecticut (UConn) in 2018. Currently, she is a postdoctoral research associate at University of Minnesota, Twin Cities. Her research interests span the areas of machine learning, data science, and network science, with special focus on Bayesian inference, Bayesian optimization, and spatio-temporal inference over graphs. In the past, she has worked on statistical signal processing, multiple target tracking, and underwater acoustic communications. She was awarded Summer Fellowship and Doctoral Dissertation Fellowship at UConn.  She was also a recipient of the Women of Innovation Award in Collegian Innovation and Leadership by Connecticut Technology Council in March, 2018.
\end{IEEEbiography}
\vspace{-1cm}
\begin{IEEEbiography}[{\includegraphics[width=1in,height=1.25in]{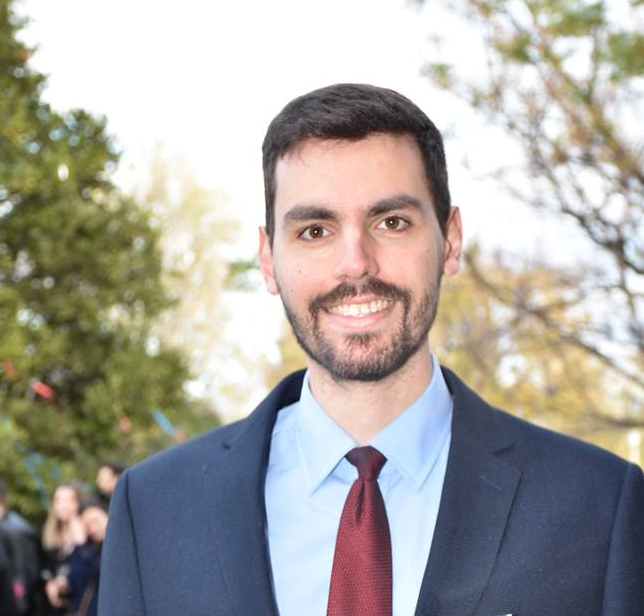}}]{Konstantinos D. Polyzos} obtained his Diploma (5-years degree) from the Department of Electrical Engineering and Computer Technology at the University of Patras, Greece in 2018. Currently, he is a third year PhD student at the Department of Electrical and Computer Engineering (ECE) at the University of Minnesota (UMN) – Twin Cities. He is a member of the SPiNCOM research group under the supervision of Prof. Georgios B. Giannakis. His research interests span the areas of machine learning, signal processing, network science and data science. Lately, he focuses on learning over graphs which can model complex networks including financial, social and biological ones to list a few. In the past, he has worked on the development of automatic aerial target recognition systems using passive Radar data. He has been awarded the UMN ECE Department fellowship (2019), Gerondelis Foundation scholarship (2020), Onassis Foundation scholarship (2021), the BEST PAPER AWARD at the International CIT$\&$DS 2019 International Conference (2019) and the Outstanding REVIEWER AWARD (top 10 \%) at the International Conference on Machine Learning (ICML 2022).
\end{IEEEbiography}
\vspace{-1cm}
\begin{IEEEbiography}[{\includegraphics[width=1in,height=1.25in]{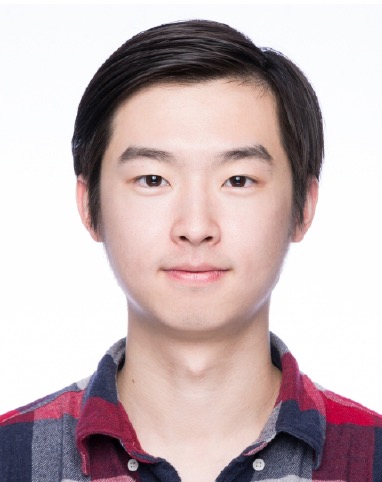}}]{Bingcong Li}
Bingcong Li received the M.Sc. and Ph.D. degree in Electrical and Computer Engineering (ECE) from the University of Minnesota (UMN), in 2019 and 2022, respectively. He is now with Huawei as a research engineer.
His research interests lie in optimization and machine learning systems. 
He received the National Scholarship twice from China in 2014 and 2015, and
UMN ECE Department Fellowship in 2017.
\end{IEEEbiography}
\vspace{-1cm}
\begin{IEEEbiography}[{\includegraphics[width=1in,height=1.25in]{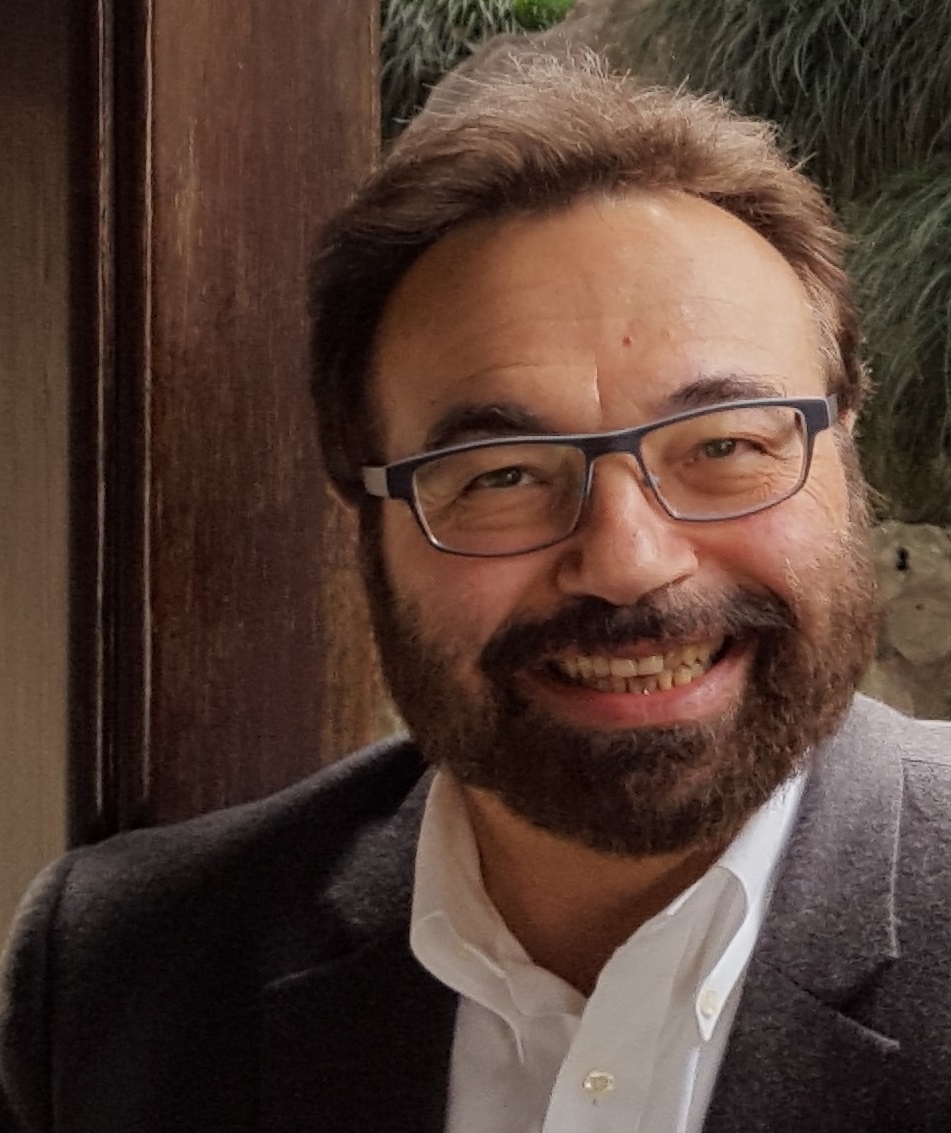}}]{Georgios~B.~Giannakis (F'97)} received his Diploma in Electrical Engr. from the Ntl. Tech. Univ. of Athens, Greece, 1981. From 1982 to 1986 he was with the Univ. of Southern California (USC), where he received his MSc. in Electrical Engineering, 1983, MSc. in Mathematics, 1986, and Ph.D. in Electrical Engr., 1986. He was a faculty member with the University of Virginia from 1987 to 1998, and since 1999 he has been a professor with the Univ. of Minnesota, where he holds an ADC Endowed Chair, a University of Minnesota McKnight Presidential Chair in ECE, and serves as director of the Digital Technology Center. His general interests span the areas of statistical learning, signal processing, communications, and networking - subjects on which he has published more than 480 journal papers, 780 conference papers, 25 book chapters, two edited books and
two research monographs. Current research focuses on Data Science,
and Network Science with applications to the Internet of Things,
and power networks with renewables. He is the (co-) inventor of
34 issued patents, and the (co-) recipient of 10 best journal
paper awards from the IEEE Signal Processing (SP) and Communications
Societies, including the G. Marconi Prize Paper Award in Wireless
Communications. He also received the IEEE-SPS Norbert Wiener
Society Award (2019); EURASIP's A. Papoulis Society Award (2020);
Technical Achievement Awards from the IEEE-SPS (2000) and from
EURASIP (2005); the IEEE ComSoc Education Award (2019); and the
IEEE Fourier Technical Field Award (2015). He is a member
of the Academia Europaea, and Fellow of the National Academy of
Inventors, the European Academy of Sciences, IEEE and EURASIP.
He has served the IEEE in a number of posts, including that
of a Distinguished Lecturer for the IEEE-SPS.
\end{IEEEbiography}

\clearpage
\appendices

\begin{figure}
	\centering
	\begin{minipage}{.495\textwidth}
		\centering
\includegraphics[width=0.6\textwidth, height=0.4\textwidth ]{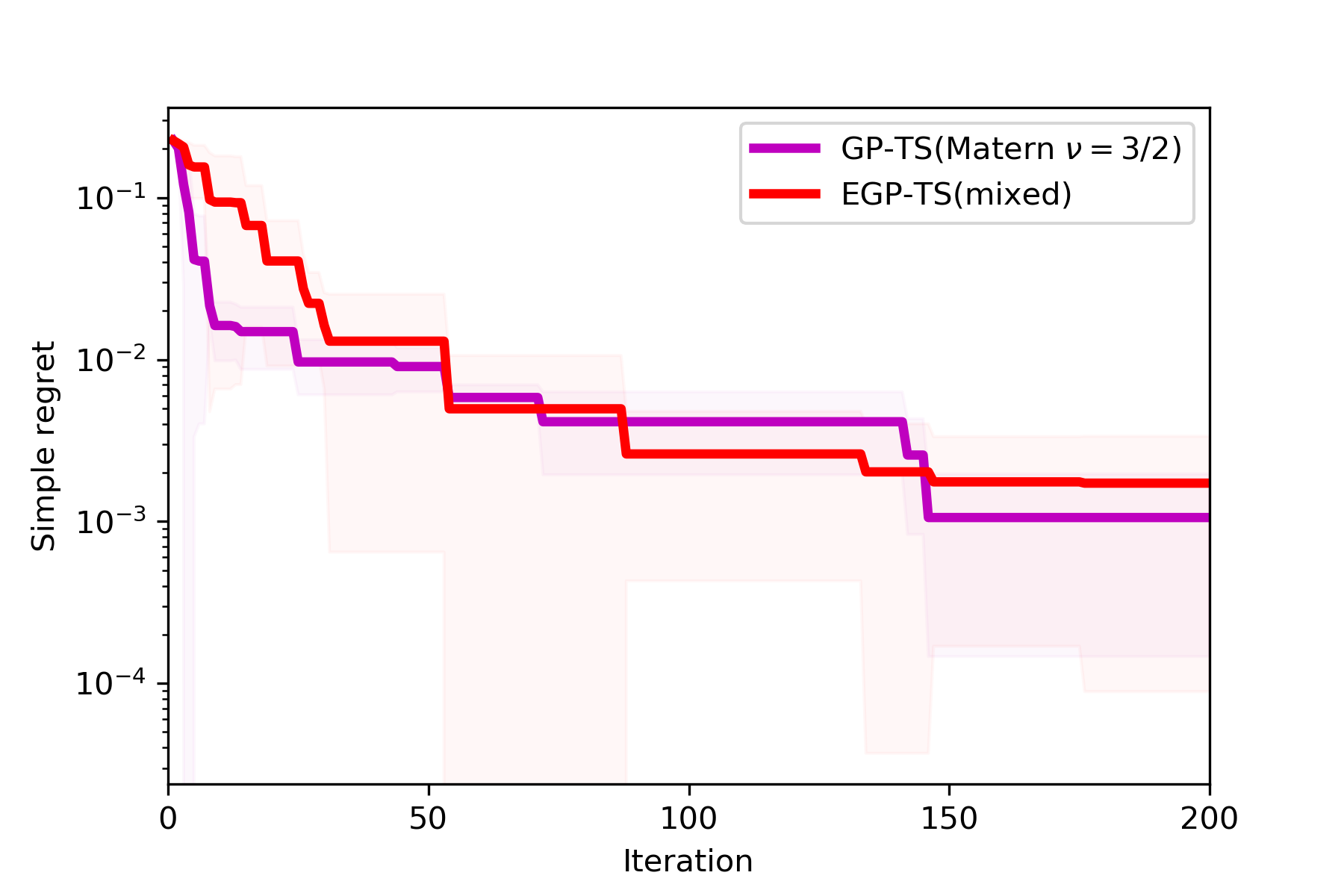}\label{robota}\\{(a)}
	\end{minipage} 
	\begin{minipage}{.495\textwidth}
		\centering
		\includegraphics[width=0.6\textwidth,height=0.4\textwidth]{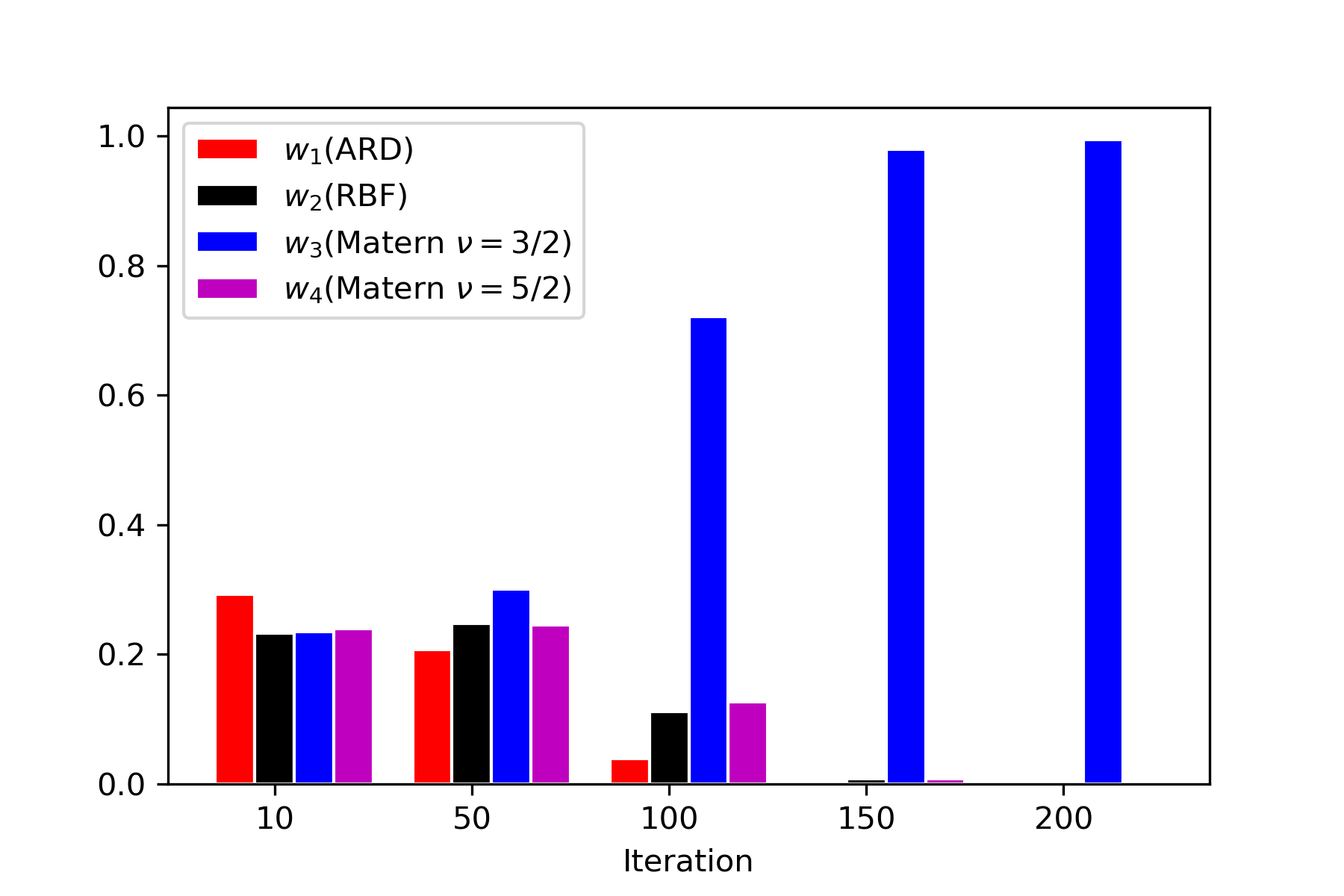}\label{robotb}\\
		{(b)}
	\end{minipage}
	\vspace*{-0.2cm}
	\caption{(a) Simple regret on black-box function drawn from a GP prior associated with a Matern kernel with $\nu=3/2$ and lengthscale equal to 1. Dictionary has 4 kernels with distinct forms: RBF with(out) ARD and Mat{\'e}rn with $\nu=3/2, 5/2$. (b) The weights of kernels in the dictionary of EGP-TS (mixed)  }
	\label{fig:priorweights}
\end{figure}

\begin{figure}
	\centering
	\begin{minipage}{.495\textwidth}
		\centering
\includegraphics[width=0.6\textwidth, height=0.4\textwidth ]{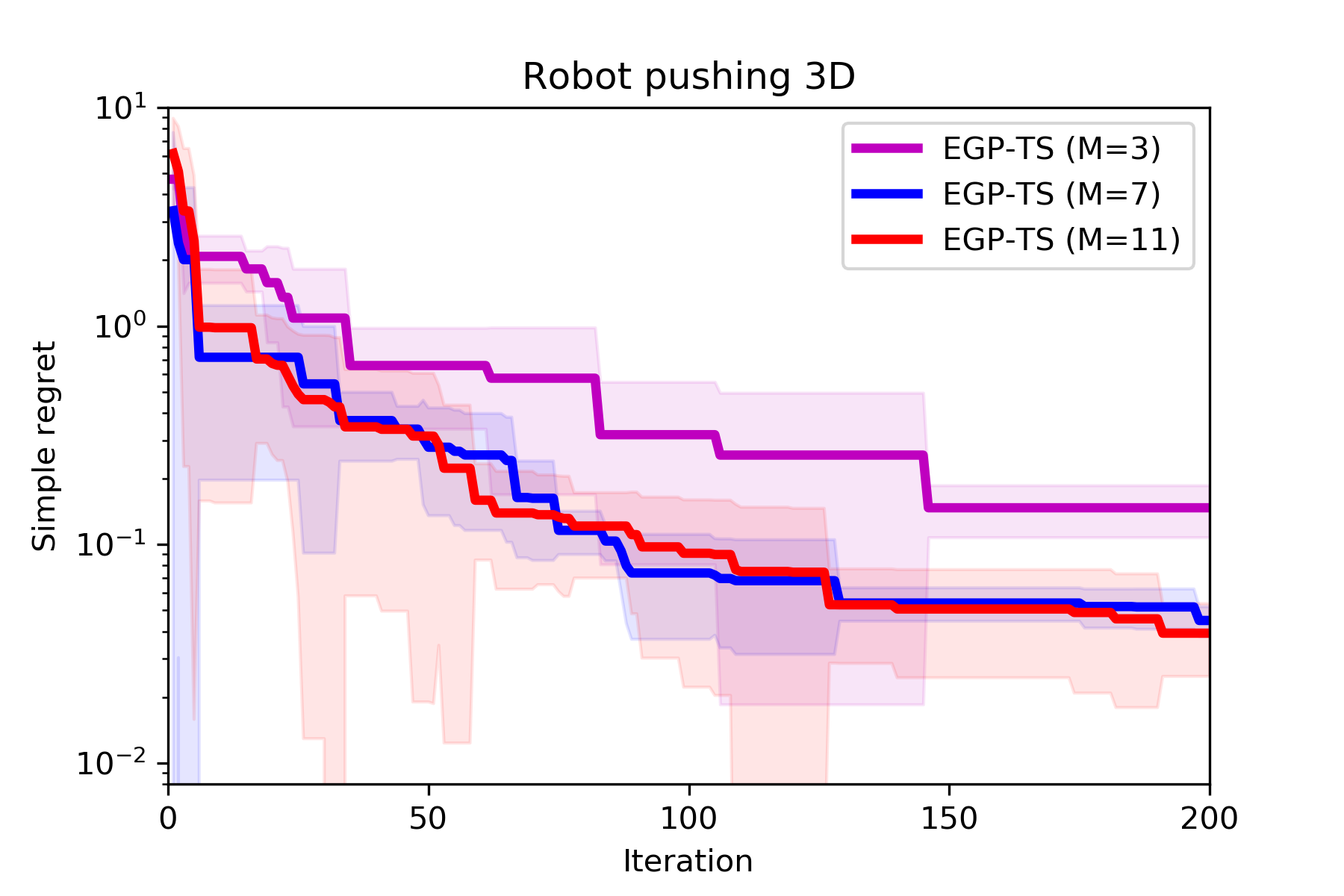}\label{robota}\\{(a)}
	\end{minipage} 
	\begin{minipage}{.495\textwidth}
		\centering
		\includegraphics[width=0.6\textwidth,height=0.4\textwidth]{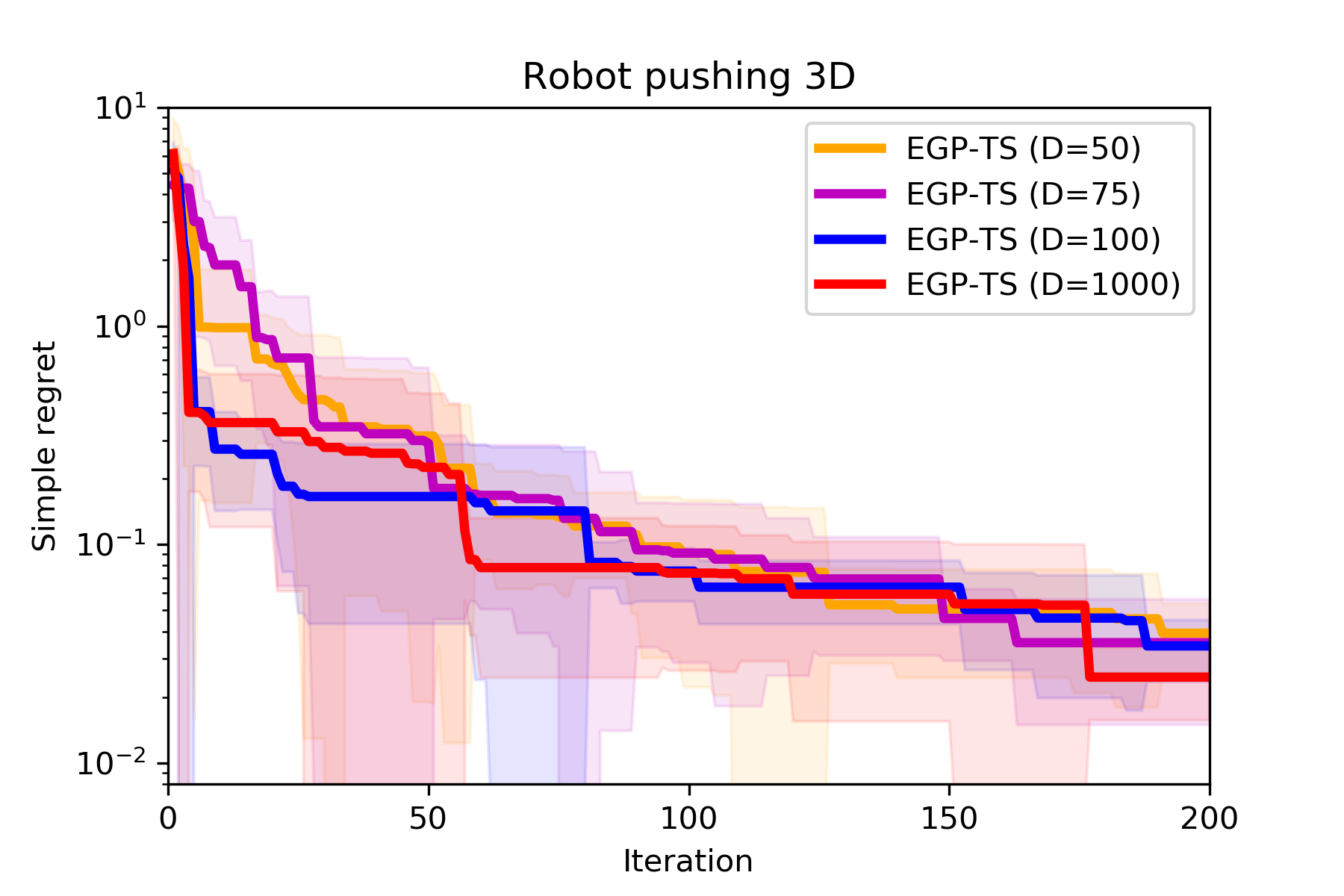}\label{robotb}\\
		{(b)}
	\end{minipage}
	\vspace*{-0.2cm}
	\caption{Simple regret on \texttt{Robot pushing 3D} for EGP-TS (a) with different number of experts $M$ (top) and (b) with different number of RFs $D$ (bottom).}
	\label{fig:ablation_M_D}
\end{figure}

\begin{algorithm}[t]
			\caption{Asynchronously parallel EGP-TS}\label{Alg: EGP-TS_asy}
			\begin{algorithmic}[1]
				\State{\textbf{Input:}  Kernel dictionary $\mathcal{K}$, number $D$ of RFs, number $K$ of workers, and $w_0^m = 1/M$ $\forall m$  }.
			\newline	\State{\textbf{Initialization:}  }
				\State Randomly evaluate $t_0$ points to obtain ${\cal D}_{t_0}$;
				\For{$m = 1, 2, \ldots, M$}
	\State Obtain kernel hyperparameters estimates $\hat{\boldsymbol{\alpha}}_{t_0}^m$ by maximizing the marginal likelihood;
	\State Draw $\nrf$ random vectors $\{\mathbf{v}_i^m\}_{i = 1}^{\nrf}$ from  $\pi_{\bar{\kappa}}^m (\mathbf{v})$ using $\hat{\boldsymbol{\alpha}}_{t_0}^m$;
		\State Obtain $w_{t_0}^m$, $\hat{\bbtheta}_{t_0}^m$, and $\bbSig_{t_0}^m$  based on~\eqref{eq:w_t} and~\eqref{eq:batch_parameter};
				\EndFor
		\For{$k = 1, 2, \ldots, K$}
		\State Sample $m_{t_0}^k$ based on pmf $\mathbf{w}_{t_0}$;
		\State Sample $\tilde{\bbtheta}_{t_0}^k$ from $\mathcal{N}(\hat{\bbtheta}^{m_{t_0}^k}_{t_0},\bbSig^{m_{t_0}^k}_{t_0})$;
		\vspace{0.1cm}
		\State Obtain $\mathbf{x}_{t_0+1}^k = \underset{\mathbf{x}\in \mathcal{X}}{\arg \max}  \ \ \tilde{\bbtheta}_{t_0}^{k\top}\!\!\bbphi^{m_{t_0}^k}(\mathbf{x})$;
		\EndFor
				\newline
				\For{$t=t_0+1, \ldots$}
				\vspace{0.1cm}
				\State Wait for a worker to obtain $\{\mathbf{x}_{t}, y_{t}\}$, and ${\cal D}_{t} = {\cal D}_{t-1}\cup \{\mathbf{x}_{t}, y_{t}\}$;	\If{Reinitialization}
		\For{$m = 1, 2, \ldots, M$}
		\State Obtain $\hat{\boldsymbol{\alpha}}_{t}^m$ by marginal likelihood maximization using ${\cal D}_t$;
		\State Draw $\nrf$ random vectors $\{\mathbf{v}_i^m\}_{i = 1}^{\nrf}$ from  $\pi_{\bar{\kappa}}^m (\mathbf{v})$ using $\hat{\boldsymbol{\alpha}}_{t}^m$;
		\State Obtain $w_{t}^m$, $\hat{\bbtheta}_{t}^m$, and $\bbSig_{t}^m$  based on~\eqref{eq:w_t} and~\eqref{eq:batch_parameter};
		\EndFor
		\Else
		\State Obtain updated $\{w_{t}^m,\hat{\bbtheta}_{t}^m,\bbSig_{t}^m \}_{m}$ using $\{\mathbf{x}_{t}, y_{t}\}$ based on~\eqref{eq:w_update} and ~\eqref{eq:theta_up};
		\EndIf
		\vspace{0.1cm}
				\State Sample $m_t$ based on pmf $\mathbf{w}_t$;
				\State Sample $\tilde{\bbtheta}_t$ from $\mathcal{N}(\hat{\bbtheta}^{m_t}_t,\bbSig^{m_t}_t)$;
				\vspace{0.1cm}
				\State Obtain $\mathbf{x}_{t+1} = \underset{\mathbf{x}\in \mathcal{X}}{\arg \max}  \ \ \tilde{\bbtheta}_t^{\top}\!\!\bbphi^{m_t}(\mathbf{x})$;
				\State Assign $\mathbf{x}_{t+1}$ to the worker to evaluate;
				\vspace{0.1cm}
				\EndFor
			\end{algorithmic}
\end{algorithm}

\section{Random feature (RF) approximation for GP}
The random feature (RF) approximation starts with a standardized shift-invariant kernel, that is,
$\bar{\kappa}(\mathbf{x}, \mathbf{x}') = \bar{\kappa}(\mathbf{x}-\mathbf{x}')$~\cite{rahimi2008random}. Bochner's theorem allows one to express any continuous $\bar{\kappa}$ as the inverse Fourier transform of a spectral density $\pi_{\bar{\kappa}}(\mathbf{v})$, expressed as~\cite{rudin1964principles}
\begin{align}
	\bar{\kappa}(\mathbf{x}-\mathbf{x}') = \int \pi_{\bar{\kappa}} (\mathbf{v}) e^{j\mathbf{v}^\top (\mathbf{x} - \mathbf{x}')} d\mathbf{v}:= \mathbb{E}_{\pi_{\bar{\kappa}}} \left[e^{j\mathbf{v}^\top (\mathbf{x} - \mathbf{x}')} \right] \nonumber
\end{align}
where the expectation in the last equality follows since $\bar{\kappa}(\mathbf{0})=\int \pi_{\bar{\kappa}}(\mathbf{v})d \mathbf{v}=1$, thus allowing one to view $\pi_{\bar{\kappa}}$ as a pdf. For instance, if $\bar{\kappa}(\mathbf{x},\mathbf{x}') = \exp(-\|\mathbf{x}- \mathbf{x}'\|^2)$, the spectral density $\pi_{\bar{\kappa}}(\mathbf{v})$ is then a standard Gaussian pdf.

\begin{figure}
	\centering
	\begin{minipage}{.495\textwidth}
		\centering
		\includegraphics[width=0.6\textwidth, height=0.4\textwidth ]{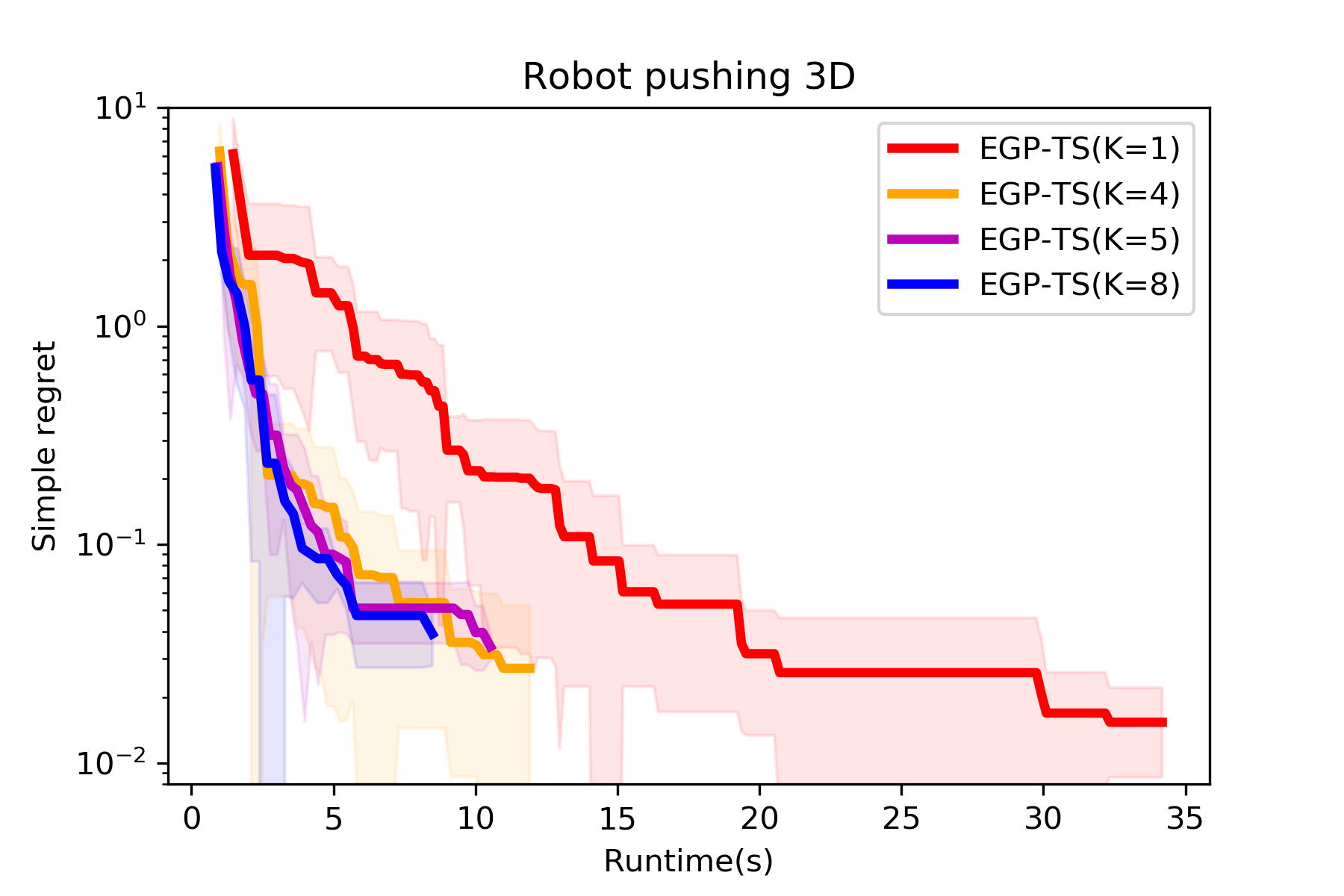}\label{robota}\\{(a)}
	\end{minipage} 
	\begin{minipage}{.495\textwidth}
		\centering
		\includegraphics[width=0.6\textwidth,height=0.4\textwidth]{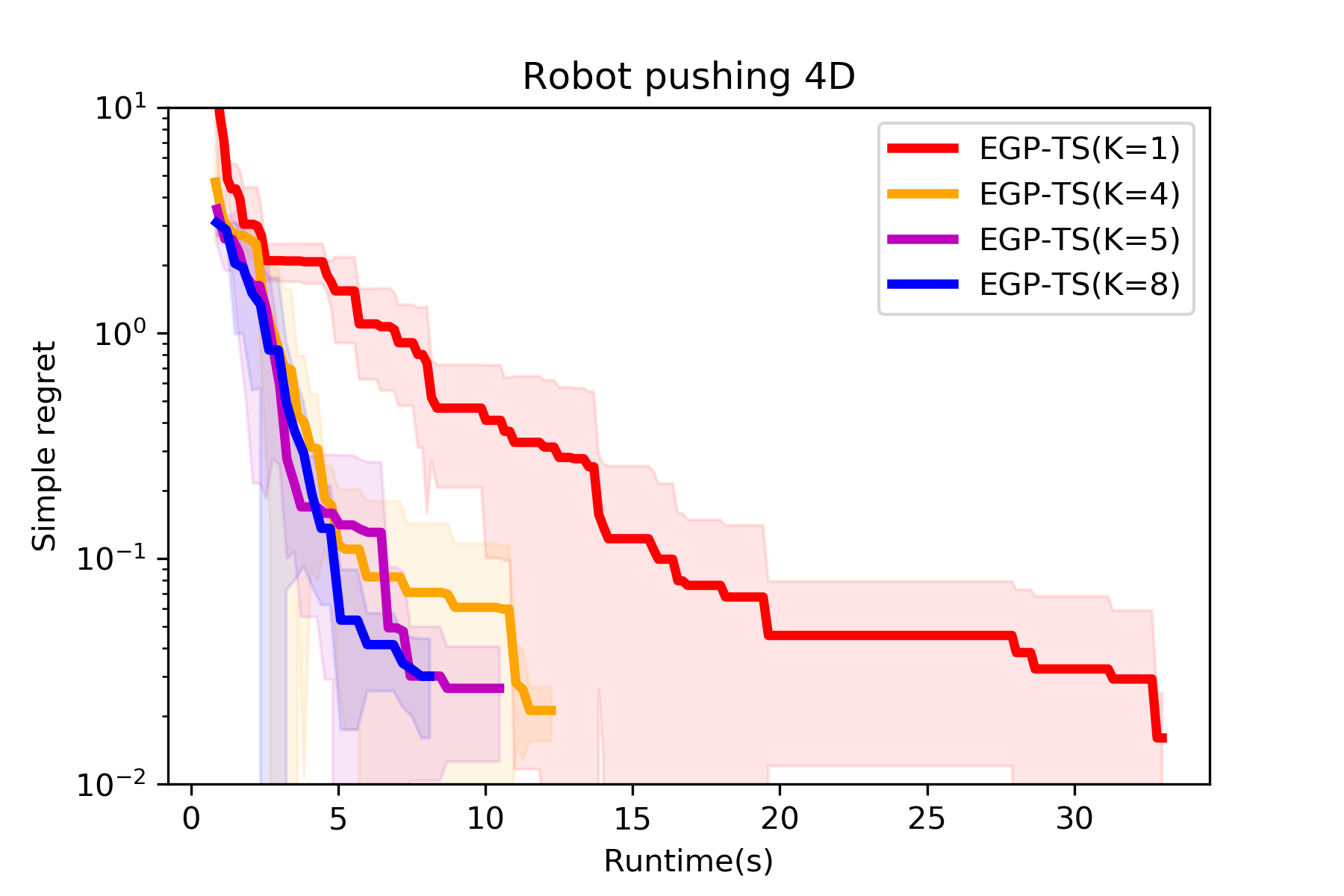}\label{robotb}\\
		{(b)}
	\end{minipage}
	\vspace*{-0.1cm}
	\caption{Simple regret vs runtime (s) on (a) \texttt{Robot pushing 3D}  and (b) \texttt{Robot pushing 4D}  for EGP-TS with mixed kernels .}
	\label{fig:ablation_parallel}
\end{figure}

Since $\bar{\kappa}$ is real, the expectation  $\mathbb{E}_{\pi_{\bar{\kappa}}}$ can be rewritten as $\mathbb{E}_{\pi_{\bar{\kappa}}} \left[\cos(\mathbf{v}^\top (\mathbf{x} - \mathbf{x}'))\right]$, which,
upon drawing a sufficient number $\nrf$ of  independent and identically distributed (i.i.d.) samples $\{\mathbf{v}_j \}_{j = 1}^{\nrf}$ from $\pi_{\bar{\kappa}} (\mathbf{v})$, can be approximated by
\begin{align}
	\check{\bar{\kappa}} (\mathbf{x}, \mathbf{x}'):= \frac{1}{\nrf} \sum_{j = 1}^{\nrf} \cos  \left(\mathbf{v}_j^\top (\mathbf{x} - \mathbf{x}') \right) \;. \label{kern_est}
\end{align}
Define the $2\nrf\!\times\! 1$ RF vector as~\cite{quia2010sparse}
\begin{align}
	&\bbphi_{\mathbf{v}} (\mathbf{x})  =\! \frac{1}{\sqrt{\nrf}}\!\left[\sin(\mathbf{v}_1^\top \mathbf{x}), \cos(\mathbf{v}_1^\top \mathbf{x}), \ldots, \sin(\mathbf{v}_{\nrf}^\top \mathbf{x}), \cos(\mathbf{v}_{\nrf}^\top \mathbf{x})\right]^{\top} \label{eq:phi_x}:
\end{align}
and use it to rewrite $\check{\bar{\kappa}}$ in \eqref{kern_est} as
$\check{\bar{\kappa}}(\mathbf{x}, \mathbf{x}') = \bbphi_{\mathbf{v}}^{\top} (\mathbf{x})\bbphi_{\mathbf{v}}(\mathbf{x}')
$. This allows approximating the nonparametric GP prior with $\kappa = \sigma_\theta^2\bar{\kappa}$ by the linear parametric one as
\begin{align}
	{\check f} (\mathbf{x}) =  \bbphi_{\mathbf{v}}^\top (\mathbf{x}) \bbtheta, \quad \bbtheta\sim \mathcal{N}(\bbtheta; \mathbf{0}_{2\nrf}, \sigma_\theta^2\mathbf{I}_{2\nrf})\;. \label{eq:f_check}
\end{align}
Henceforth, the function posterior pdf can be captured by $p(\bbtheta|\mathcal{D}_t) = \mathcal{N}(\bbtheta; \hat{\bbtheta}_t, \bbSig_t)$, where the first two moments are given by ($\bbPhi_t:=\left[\bbphi_{\mathbf{v}}(\mathbf{x}_1), \ldots, \bbphi_{\mathbf{v}}(\mathbf{x}_t) \right]^\top$)
\begin{subequations}
\begin{align}
\hat{\bbtheta}_t & = \left( \bbPhi_t^{\top} \bbPhi_t+ \frac{\noisevar}{\sigma_\theta^2}\mathbf{I}_{2D} \right)^{-1}\!\!\!\! \bbPhi_t^\top \mathbf{y}_t  \\
\bbSig_t & = \left(\!\!\frac{\bbPhi_t^{\top} \bbPhi_t}{\noisevar} \!+\!  \frac{\mathbf{I}_{2D}}{\sigma_\theta^2} \!\!\right)^{-1} \;.
\end{align}\label{eq:batch_parameter}
\end{subequations}

\begin{figure*}[t]
	\centering
	\includegraphics[width=1\linewidth]{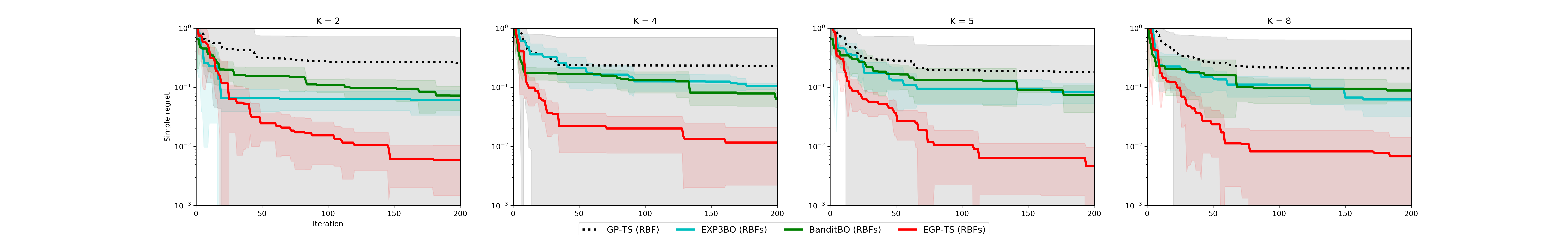}
	\vspace*{-0.2cm}
	\caption{Simple regret for the parallel setting for various $K$s on \texttt{Ackley-5d} function. Dictionary has 11 RBF kernels with characteristic lengthscales given by $\{10^c\}_{c=-4}^6$.} \label{fig:SR_parallel_ackley}
\end{figure*}

\section{Additional experimental results}
\subsection{Ablation studies}
Several ablation studies are presented here to further demonstrate the performance of the proposed EGP-TS approach.
\subsubsection{Function sampled from a GP prior }
We conducted an experiment in which the black-box function is a draw from a GP prior associated with a Matern kernel with $\nu = 3/2$ and lengthscale equal to 1. The performance of the advocated EGP-TS with $4$ mixed kernels is compared against the GP-TS counterpart with the same Matern kernel that is used to generate the black-box function. Fig.~12 (a) shows the simple regret performance of the two competing approaches. As expected, the EGP-TS(mixed) approach has slightly inferior but comparable performance relative to GP-TS (Matern $\nu = 3/2$). The weights of the different kernels in the dictionary of EGP-TS(mixed) are further depicted in Fig.~12 (b), where it is evident that the weight of GP expert with Matern kernel with $\nu = 3/2$ and lengthscale 1 ($w_3$) goes to $1$ as more data are collected. This test validates that the proposed EGP-TS 
can indeed select the correct GP prior from the ensemble.

\subsubsection{ Effect of the number $M$ of kernels} Selecting the number of kernels ($M$) in the dictionary to attain a desirable performance-complexity trade-off depends on the problem at hand and the available level of domain knowledge. If the sought posterior is known to have more than $M$ modes, intuition suggests that improvements will saturate if EGP comprises more than $M$ components. We conducted an ablation study to ascertain how the performance of EGP-TS (dictionary has 11 RBF kernels with characteristic lengthscales given by $\{10^c\}_{c=-4}^6$) changes with the number $M$ of models in the dictionary on the \texttt{Robot pushing 3D} task. As shown Fig.~13(a), increasing $M$ indeed yields improved performance. But the improvement gap grows smaller when $M$ goes from $3$ to $7$.
 

\subsubsection{ Effect of the number $D$ of spectral features}
An ablation study has been conducted to see how the performance of EGP-TS changes with $D$ on \texttt{Robot pushing 3D} task. As shown in Fig.~13 (b), the performance gain is negligible when $D$ goes from $50$ to $1000$.

\subsection{Results of parallel EGP-TS}
The regret performance of synchronous parallel EGP-TS is plotted as a function of the running time in Fig.~14 for $K = 1,4,5,8$ on the \texttt{Robot pushing 3D and 4D} tasks. It is evident that increasing $K$ yields reduced running time with slightly larger regret performance (in accordance with Theorem 3). In addition, Fig.~\ref{fig:SR_parallel_ackley} presents the results of the competing parallel TS-based methods in the synchronous mode for various $K$s on the \texttt{Ackley-5d} function. Clearly, EGP-TS also stands out from the rest of the baselines in this batch setting.  

\subsection{Average runtimes on synthetic functions and robotic tasks}
Table.~\ref{table:allruntimes} contains the average runtimes of competing methods on synthetic functions and robotic tasks (The fully Bayesian GP method was not reported since it requires a much larger runtime due to the time-consuming MCMC step). It can be observed that the proposed EGP-TS requires slightly increased runtime relative to the single GP-TS alternatives.

\begin{table*}
	\caption{Average runtimes of competing methods on synthetic functions and robotic tasks} \label{table:allruntimes}
	\begin{center}
		
		\begin{tabular}{c|c|c|c|c|c|c}
			\hline
			\hline
                  & Ackley5D & Zakharov & Drop-wave & Eggholder& Robot pushing 3D & Robot pushing 4D \\
                \hline
                GP-TS(RBF) & 30.9906 & 34.6408 & 12.4251 & 27.9748 & 29.5710 & 31.5313 \\
                \hline
                GP-TS (Matern $\nu=5/2$) & 70.5168 & 51.9291 & 13.4455 & 29.5407 & 29.3256 & 32.4717 \\
                \hline
                GP-TS (Matern $\nu=3/2$) & 64.5795 & 48.3568 & 14.1249 & 28.4913 & 29.7532 & 31.2688
                \\
                \hline
                GP-TS(ARD) & 44.8131 & 33.0074 & 30.1294 & 30.0869 & 31.3012 & 31.8359 \\
                \hline
                GP-EI & 128.8359 & 244.3411 & 82.7003 & 110.0964 & 86.3489 & 168.9366 \\
                \hline
                EGP-TS(mixed) & 41.0102 & 41.9205 & 19.0502 & 33.6314 & 34.1852 & 32.9830 \\
                \hline
                EGP-TS(RBFs) & 40.3958 & 48.1003 & 19.9600 & 35.4065 & 31.3028 & 33.3244 \\
                \hline
                BanditBO(RBFs) & 334.6085 & 374.4101 & 184.5911 & 322.8238 & 329.5755 & 376.3908 \\
                \hline
                BanditBO(mixed) & 141.1233 & 145.0351 & 51.7992 & 116.6906 & 109.8294 & 120.1719 \\
                \hline
                EXP3BO(RBFs) & 36.2382 & 49.4687 & 18.3948 & 27.2058 & 29.4610 & 33.4458 \\
                \hline
                EXP3BO(mixed) & 31.9941 & 36.5181 & 12.7743 & 31.1013 & 29.4353 & 31.9681\\
			\hline
		\end{tabular}
	\end{center}
\end{table*}

\begin{table*}
	\caption{Candidate values of $\mathbf{x}$, maximizer $\mathbf{x}_*$, and $f(\mathbf{x}_*)$ for the synthetic functions} \label{table:stat_synth}
	\vspace{0.1cm}
	\begin{center}
		
		\begin{tabular}{c|c|c|c|c}
			\hline
			\hline
			Synthetic function & Lower bound & Upper bound  & ${\bf x}_*$ &	$f(\bf x_*)$  \\
			\hline
			\hline 
			Ackley-5d 	 & [0, 0, 0, 0, 0] & [1, 1, 1, 1, 1]	&  [0.6231,0.6231,1,0.6231,0.6231] &  4.6930  \\
			\hline 		
			Zakharov 	&  [-5, -5, -5, -5] & [10, 10, 10, 10] & [0,0,0,0] & 0	   \\
			\hline 
			Drop-wave	&  [-5.12, -5.12] & [5.12, 5.12]  & [0,0] & 1 \\
			\hline
			Eggholder	&  [-512, -512] & [512, 512]&  [512, 404.2319] & 959.6407  \\
			\hline
		\end{tabular}
	\end{center}
\end{table*}

\subsection{Additional information about synthetic functions}. 
The analytical expressions of the evaluated synthetic functions are as follows:
\begin{itemize}
	\item {Ackley-5d}: $ f({\bf x})=-20 \exp(-0.2(\sum_{i=1}^5 x_i^2/5)^{1/2})  -\exp(\sum_{i=1}^5 \cos(2\pi x_i)/5)  + 20 + e^{1}$	
	\item Zakharov: $f({\bf x}) = -\sum_{i=1}^d x_i^2 - \left(\sum_{i=1}^d 0.5 i x_i\right)^2 - \left(\sum_{i=1}^d 0.5 i x_i\right)^4$
	\item Drop-wave: $ f({\bf x}) = 
	(1+\cos (12\sqrt{x_1^2+x_2^2}))/(0.5 (x_1^2+x_2^2)+2)$
	\item Eggholder: $f({\bf x}) = (x_2 +47)\sin \sqrt{|x_2+x_1/2+47|}+x_1 \sin \sqrt{|x_1 -x_2 -47|}$
\end{itemize}
where the candidate values of $\mathbf{x}$, the corresponding optimizer $\mathbf{x}_*$, as well as maximum function value $f(\mathbf{x}_*)$ are provided in Table~\ref{table:stat_synth}. Note that the latter three functions are the opposite of the standard forms in order to transform the minimization problem to a maximization one.

\end{document}